\documentclass[10pt,twocolumn,letterpaper]{article}

\usepackage{cvpr}

\usepackage{url}            
\usepackage{booktabs}       
\usepackage{amsfonts}       
\usepackage{nicefrac}       
\usepackage{microtype}      
\usepackage{color,xcolor,colortbl}   
\usepackage{graphicx}
\usepackage[accsupp]{axessibility}  

\usepackage{amsmath}
\usepackage{amssymb}
\usepackage{bm}
\usepackage{booktabs}
\usepackage{epsfig}
\usepackage{epigraph}
\usepackage{threeparttable}
\usepackage{tabularx}
\usepackage{tabularray}
\usepackage{wrapfig}
\usepackage{multirow}
\usepackage{float}
\usepackage{overpic}
\usepackage{appendix}
\usepackage{float}
\usepackage{booktabs} 
\usepackage{diagbox}
\usepackage{makecell}
\usepackage{bbm}
\usepackage{dsfont}
\usepackage{cuted}
\usepackage{arydshln}
\usepackage{overpic}
\usepackage[misc]{ifsym}
\usepackage{wrapfig}
\usepackage{xspace}
\usepackage{varwidth}
\usepackage{pifont}
\usepackage{tablefootnote}
\newcommand{\cmark}{\ding{51}}%
\newcommand{\xmark}{\ding{55}}%
\definecolor{color3}{rgb}{0.95,0.95,0.95}

\usepackage{capt-of}
\usepackage{algorithm}
\usepackage{algpseudocode}
\usepackage{etoc}
\usepackage{gensymb}
\usepackage[accsupp]{axessibility}

\definecolor{cvprblue}{rgb}{0.21,0.49,0.74}

\usepackage[pagebackref,breaklinks,colorlinks,allcolors=cvprblue]{hyperref}

\def\confName{CVPR}
\def\confYear{2025}

\title{InterAct: Advancing Large-Scale Versatile 3D Human-Object Interaction Generation}

\author{Sirui Xu$^{\dag}$ \quad Dongting Li$^{\dag}$ \quad Yucheng Zhang$^{\dag}$ \quad Xiyan Xu$^{\dag}$ \quad Qi Long$^{\dag}$ \quad Ziyin Wang$^{\dag}$ \\ Yunzhi Lu \quad Shuchang Dong \quad Hezi Jiang \quad Akshat Gupta \quad
Yu-Xiong Wang$^{\ddag}$ \quad
Liang-Yan Gui$^{\ddag}$\\
University of Illinois Urbana-Champaign \\
$^{\dag}$ Equal Contribution \quad $^{\ddag}$ Equal Advising\\
{\small{\url{https://sirui-xu.github.io/InterAct/}}}}
\begin{document}
\maketitle
\begin{strip}\centering
\vspace{-4em}
\includegraphics[width=\textwidth]{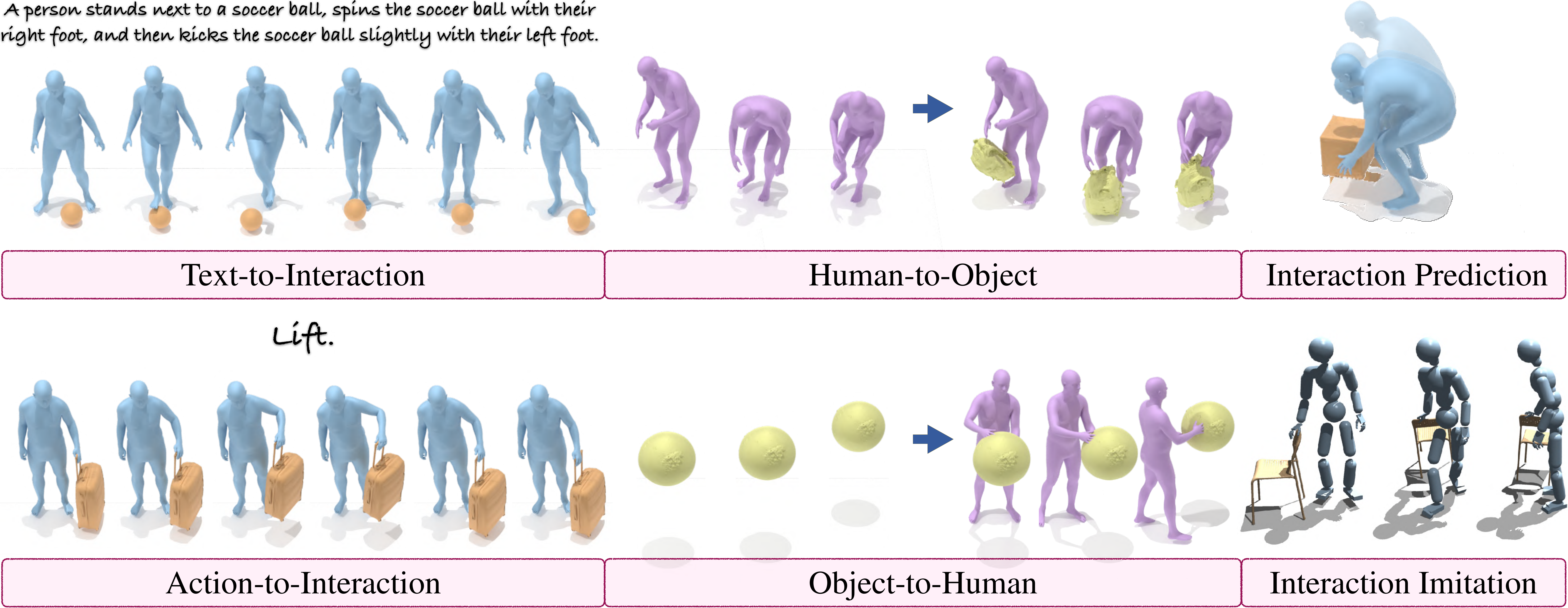}
\captionof{figure}{An overview of InterAct, our large-scale 3D human-object interaction (HOI) benchmark, covering six HOI generation tasks.
\label{fig:teaser}}
\end{strip}
\begin{abstract}
While large-scale human motion capture datasets have advanced human motion generation, modeling and generating dynamic 3D human-object interactions (HOIs) remain challenging due to dataset limitations. Existing datasets often lack extensive, high-quality motion and annotation and exhibit artifacts such as contact penetration, floating, and incorrect hand motions. To address these issues, we introduce InterAct, a large-scale 3D HOI benchmark featuring dataset and methodological advancements. First, we consolidate and standardize 21.81 hours of HOI data from diverse sources, enriching it with detailed textual annotations. Second, we propose a unified optimization framework to enhance data quality by reducing artifacts and correcting hand motions. Leveraging the principle of contact invariance, we maintain human-object relationships while introducing motion variations, expanding the dataset to 30.70 hours. Third, we define six benchmarking tasks and develop a unified HOI generative modeling perspective, achieving state-of-the-art performance. Extensive experiments validate the utility of our dataset as a foundational resource for advancing 3D human-object interaction generation. To support continued research in this area, the dataset is publicly available at \url{https://github.com/wzyabcas/InterAct}, and will be actively maintained.

\end{abstract}
\section{Introduction}
Recent advances in human motion modeling have significantly benefited from extensive motion capture (MoCap) datasets~\cite{mahmood2019amass,BABEL:CVPR:2021,guo2022generating,lin2023motionx,plappert2016kit}, enabling the creation of scalable generative models for diverse human movements. Building upon this foundation, researchers are increasingly turning to the more intricate challenge of generating human-object interactions (HOIs)~\cite{xu2023interdiff,xu2024interdreamer,xu2025intermimic}. This emerging area holds considerable promise for applications in robotics, animation, and computer vision.

However, high-quality HOI generation faces notable obstacles due to factors such as increased degrees of freedom introduced by objects, varied object geometries, dynamic interactions, and the necessity for physically accurate contact modeling. Current methods often struggle to achieve realism primarily because existing datasets lack scalability and comprehensive annotations, which are crucial for models to effectively understand interaction dynamics and link them to related domains such as natural language.

Specifically, these challenges underscore the need for comprehensive, high-quality HOI datasets:
(\textbf{1}) \textit{Limited and Inconsistent Datasets}: Existing methods typically depend on small datasets with limited hours of data, difficult to consolidate due to inconsistent human representations, object types, coordinate systems, and annotations. Available annotations~\cite{peng2023hoi,li2023object} are frequently coarse and incomplete, lacking detailed descriptions of human states, object interactions, and involved body parts.
(\textbf{2}) \textit{Prevalent Artifacts}: Current datasets often contain artifacts from MoCap limitations and occlusions, including unnatural penetrations, floating contacts, inaccurate hand poses~\cite{bhatnagar22behave,li2023object}, and significant motion jitter~\cite{huang2022intercap}. These issues compromise the models' capacity to learn realistic human-object dynamics.
\begin{table}
    \centering
    \resizebox{\linewidth}{!}{
        \makeatletter\def\@captype{table}\makeatother
        \begin{tabular}{@{}lccccc@{}}
        \toprule
        Dataset&Clip&Hour&Text&Hand&Object\\
        \midrule
        GRAB~\cite{taheri2020grab}&1,335&3.76&\xmark&\cmark&51\\
        BEHAVE~\cite{bhatnagar22behave}&299&4.13&\xmark&\xmark&18\\
        InterCap~\cite{huang2022intercap}&233&0.62&\xmark&\cmark&10\\
        Chairs~\cite{jiang2022chairs}&1,041&2.37&\xmark&\cmark&92\\
        HODome~\cite{zhang2023neuraldome}&176&2.82&\xmark&\cmark&21\\
        OMOMO~\cite{li2023object}&4,838&8.27&4,838&\xmark&15\\
        IMHD~\cite{zhao2023im}&164&0.97&\xmark&\cmark&10\\
        \midrule
        InterAct (\textbf{Ours})&\underline{11,350}&\underline{21.81}&\underline{34,050}&\cmark&\textbf{217}\\
        InterAct-X (\textbf{Ours})&\textbf{16,201}&\textbf{30.70}&\textbf{48,630}&\cmark&\textbf{217}\\ \bottomrule
            \end{tabular}
        }
        \caption{Comparison between InterAct, InterAct-X, and human-object interaction datasets we collect. Beyond a substantially larger scale, our dataset introduces comprehensive textual annotations and enhances interaction quality, offering a more versatile foundation for large-scale HOI generation.}
        \label{tab:datasets}
        \vspace{-0.5em}
\end{table}

To address these challenges, we present \textbf{InterAct}, a benchmark designed to systematically overcome current limitations and drive advancements in 3D HOI modeling. As shown in Table~\ref{tab:datasets}, InterAct offers a large-scale, standardized dataset of carefully curated interactions from existing resources\footnote{We integrate publicly available datasets~\cite{taheri2020grab, bhatnagar22behave, huang2022intercap, jiang2022chairs, zhang2023neuraldome, li2023object, zhao2023im} focusing on single-human interactions with rigid and dynamic objects. Certain datasets were selectively included based on relevance and data type.}, enriched by detailed textual annotations.

\begin{figure}
    \centering
    \includegraphics[width=\columnwidth]{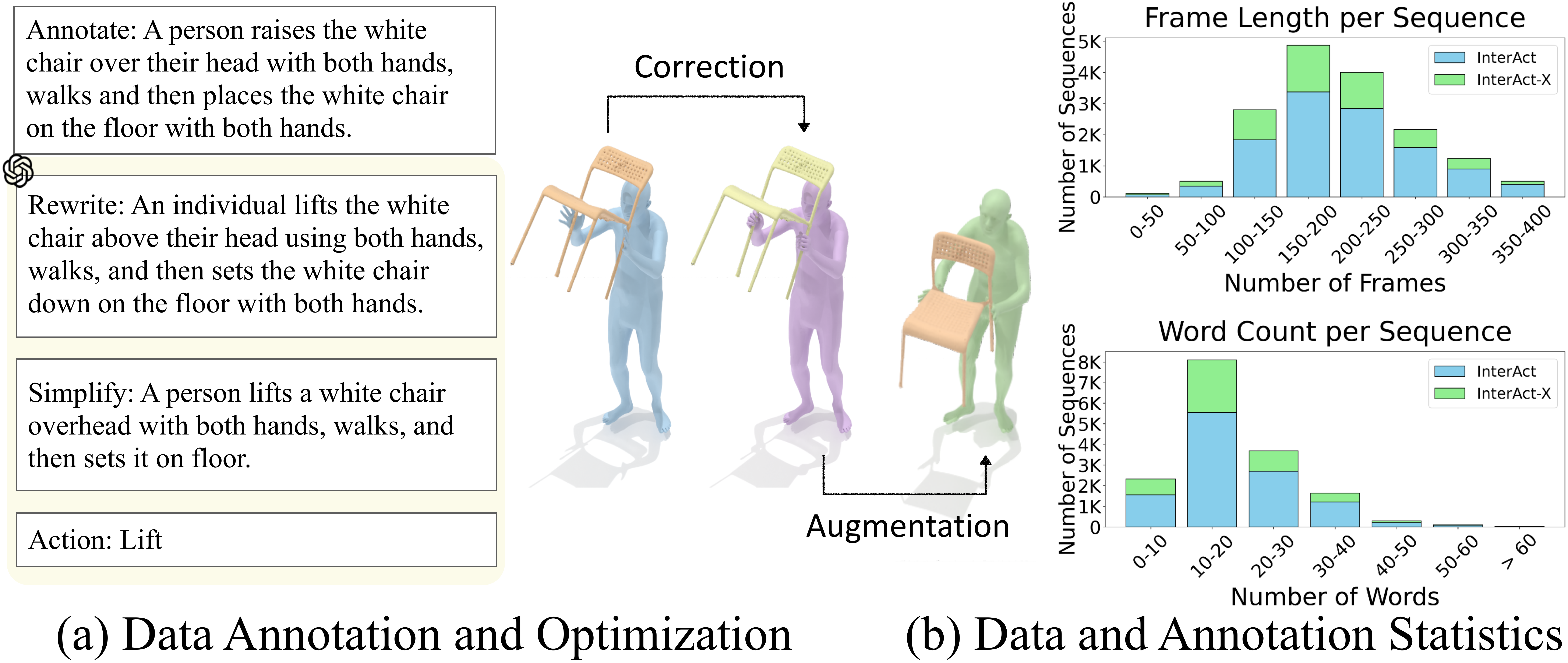}
    \caption{(a) Our data processing pipeline consolidating data, annotations via foundation models, corrections, and interaction illustrations. (b) Statistics on motion and text annotations.}
    \label{fig:teaser_info}
    \vspace{-0.5em}
\end{figure}

To further enhance dataset quality and scope, we introduce a \textit{unified optimization approach}, addressing major penetration and floating artifacts first in whole-body interactions, followed by refined corrections for nuanced hand-object interactions. Additionally, we propose the concept of \textit{contact invariance}, inspired by motion mirroring techniques, to generate realistic synthetic data by varying human motions while maintaining consistent object contacts. This augmentation expands InterAct into \textbf{InterAct-X}, providing approximately 9 additional hours of data and substantially improving generative model performance.

Leveraging this comprehensive, richly annotated dataset, we define benchmarks across six key HOI generation tasks, Text-to-Interaction, Action-to-Interaction, Object-to-Human, Human-to-Object, Interaction Prediction, and Interaction Imitation, as shown in Figure~\ref{fig:teaser}, and propose a unified modeling and representation for kinematic generative tasks. Our method utilizes multi-task learning to jointly model motion and contact, achieving state-of-the-art performance as validated by comprehensive evaluations.

In summary, our contributions are:
(\textbf{1}) \textbf{InterAct}, the most extensive 3D HOI benchmark to date, facilitating large-scale generative modeling.
(\textbf{2}) A unified optimization-based framework to correct and augment MoCap data, addressing common artifacts and significantly enhancing dataset quality. We believe this will aid future research in overcoming data scarcity before capturing potentially imperfect data. 
(\textbf{3}) Comprehensive benchmarks across six HOI generation tasks, establishing standardized metrics and demonstrating superior performance over existing approaches. This benchmark lays a strong foundation for future research, encouraging advancements across multiple facets of 3D HOI generation.
\section{Related Work}
\label{sec:related}

\noindent{\bf Dynamic 3D HOI Dataset.}
Many large-scale datasets with sequential human motion data have established benchmarks for the task of 3D human motion generation. 
However, human actions are influenced not only by individual intents but also by interactions with the surrounding environment. To address this complexity, new datasets have been developed to capture the dynamics between humans and their environments, including interactions with other humans~\cite{liang2023intergen,CMU-Mocap,singleshotmultiperson2018,xu2023inter} and scenes~\cite{hassan_samp_2021,cao2020long,hassan2019resolving}. 
Though there are fruitful hand-object interaction datasets~\cite{liu2022hoi4d,moon2020interhand2,zhan2024oakink2,ohkawa2023assemblyhands,chao2021dexycb,hampali2020honnotate,wiederhold2024hoh,zhang2024core4d,zhang2024force,lv2025himo,wiederhold2024hoh,tendulkar2023flex}, our focus is specifically on \emph{whole-body interactions with dynamic objects}, ranging from low-dynamic interactions such as approaching and manipulation~\cite{taheri2020grab} to highly dynamic interactions involving multiple body parts~\cite{bhatnagar22behave,jiang2022chairs,huang2022intercap,zhang2023neuraldome,fan2023arctic,li2023object,zhao2023im,kim2024parahome,jiang2024scaling,yang2024f}.

We aim to address the limitations of these datasets and open new possibilities for future research. Our InterAct dataset maintains advantages in motion quality, fine-grained textual annotations, detailed hand gestures, and comprehensive annotation modalities. We provide quantitative comparisons of InterAct and existing datasets in Table~\ref{tab:datasets}, demonstrating the superiority of our dataset in these aspects.

\noindent{\bf Dynamic 3D HOI Generation.}
Existing human-object interaction (HOI) datasets have laid a robust foundation for generating dynamic, whole-body interactions. Extensive research has explored the generation of hand-object interactions~\cite{li2023task,ye2023affordance,zheng2023cams,zhou2022toch,zhang2024manidext,zhang2023artigrasp,tian2024gaze,ma2024diff,liu2023contactgen,christen2024diffh2o,cha2024text2hoi} and static human-object interactions~\cite{xie2022chore,zhang2020perceiving,wang2022reconstructing,petrov2023object,hou2023compositional,kim2023ncho,yang2024lemon,yang2024person,xie2024intertrack,zhao2022compositional}. Meanwhile, full-body dynamic interactions have also been studied extensively~\cite{starke2019neural,starke2020local,taheri2022goal,wu2022saga,kulkarni2023nifty,zhang2022couch,lee2023locomotion,xu2021d3dhoi,corona2020context,9714029,razali2023action,Mandery2015a,Mandery2016b,krebs2021kit,ghosh2022imos,li2023object,Zhao:ICCV:2023}, though these often face significant limitations, including narrow action repertoires and dependence on static objects.
Recent advancements, such as InterDiff~\cite{xu2023interdiff}, have introduced diverse interactions involving dynamic objects and multiple body parts. Building upon this, subsequent approaches like InterDreamer~\cite{xu2024interdreamer} and other contemporary studies~\cite{peng2023hoi,diller2023cg,li2023controllable,wu2024thor,wu2024human,song2024hoianimator,xu2024interdreamer,zhang2024hoi} further demonstrate the feasibility of converting textual descriptions into realistic 3D human-object interaction sequences. Despite these advances, current methods remain constrained by a shortage of high-quality, large-scale datasets, often encountering issues related to physical inaccuracies, such as floating contacts or interpenetration.
In parallel, physics-based methods leveraging deep reinforcement learning (RL)~\cite{liu2018learning,chao2021learning,merel2020catch,hassan2023synthesizing,bae2023pmp,yang2022learning,xie2023hierarchical,pan2023synthesizing,braun2023physically,wang2024skillmimic,yao2024moconvq,xie2023omnicontrol,wu2023daydreamer,wang2023intercontrol,wang2024strategy,tevet2024closd,tessler2024maskedmimic,cui2024anyskill} successfully generate physically accurate interactions, with applications in sports~\cite{luo2024smplolympics} like basketball~\cite{wang2023physhoi,wang2024skillmimic} and soccer~\cite{xie2022learning}. Nonetheless, these methods typically produce rigid interaction patterns from limited datasets. InterMimic~\cite{xu2025intermimic}, instead, illustrates that physics-based approaches can digest effectively across diverse and dynamic object interactions.
Our work addresses these fundamental limitations at the dataset level by providing enhanced diversity and comprehensive sequences of human-object interactions. This facilitates multiple generative tasks, supports better contact modeling, and improves the capability to synthesize realistic and generalized human-object interactions.

\section{InterAct Dataset}
\noindent\textbf{Overview.} We introduce InterAct, the first \textit{unified} benchmark tailored explicitly for sequential 3D human-object interaction (HOI) generative modeling. Distinguished by its unprecedented \textit{scale} and \textit{comprehensiveness}, InterAct significantly surpasses existing datasets, as summarized in Table~\ref{tab:datasets}. InterAct is available in two versions:
(\textbf{1}) A \textit{basic} version consolidating seven existing datasets, providing 21.81 hours of annotated 3D whole-body interactions with corresponding semantic descriptions.
(\textbf{2}) An \textit{advanced} version, InterAct-X, extending the basic dataset through synthetic data generated via our unified optimization framework.
Figure~\ref{fig:teaser_info} presents examples of motion and text annotations. To ensure high-quality standards, we employ a multifaceted annotation strategy combining \textit{human expertise}, \textit{automated foundation models}, and \textit{advanced HOI modeling techniques}, all validated through rigorous manual quality checks.

\subsection{Data Collection, Annotation, and Unification}\label{sec:collection}

\begin{figure}
    \centering
    \includegraphics[width=\columnwidth]{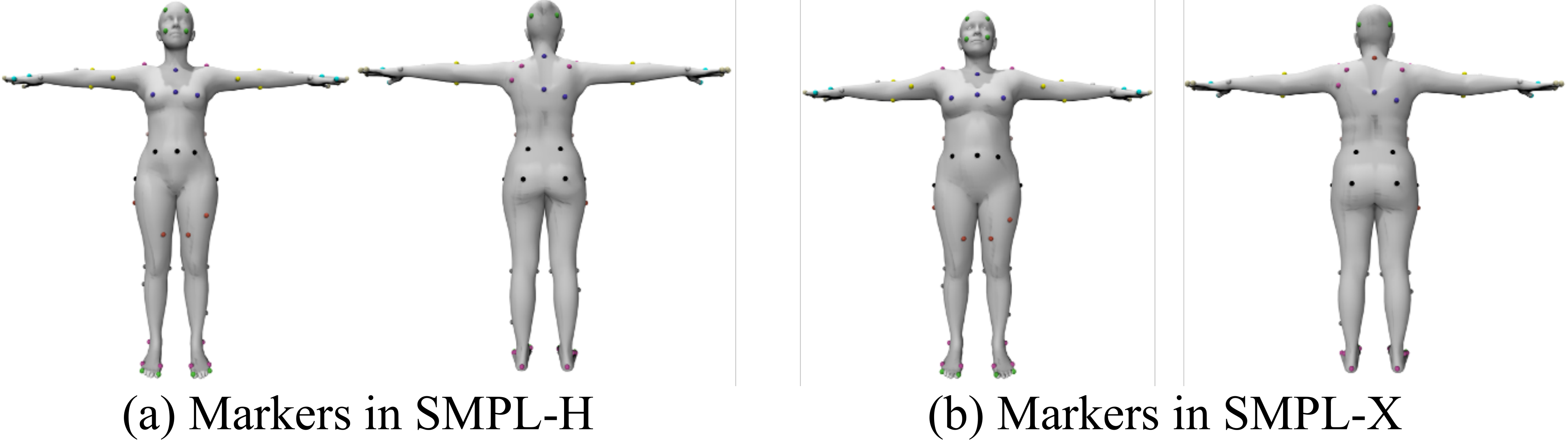}
    \caption{Marker-based representation for human.}
    \label{fig:marker}
    \vspace{-0.5em}
\end{figure}

We compile data from seven datasets~\cite{bhatnagar22behave, huang2022intercap, taheri2020grab, jiang2022chairs, zhang2023neuraldome, li2023object, zhao2023im}, featuring motion capture of a single human interacting with a single dynamic 3D object, where humans are annotated with SMPL~\cite{SMPL-X:2019,MANO,loper2015smpl}. We address their \emph{heterogeneity} in two key aspects: annotations and representations.

\noindent\textbf{Unifying Textual Annotations.} Since most datasets either lack textual descriptions or provide only very coarse text descriptions~\cite{li2023object}, we implement a \emph{two-phase} annotation procedure involving human annotators and GPT-4~\cite{chatgpt} to generate consistent and detailed annotations across all subsets.
In the \emph{first phase}, human annotators provided detailed and precise descriptions of the interactions, adhering to the following guidelines:
(\textbf{i}) Split motion sequences into clips averaging 300 frames (approximately 10 seconds) but no longer than 400 frames each;
(\textbf{ii}) Clearly describe the actions and the body parts involved in the interactions.
For example, a typical annotation is: \emph{``A person sits on a stool and touches the ground with their left hand, then their right hand.''} For the subset derived from OMOMO~\cite{li2023object}, we skip this phase and directly utilize their annotations.
In the \emph{second phase}, we use GPT-4 to rephrase and simplify human annotations to enhance diversity and consistency. For example, the rephrased version is \emph{``A person perches on a stool, touching the ground with their left hand, then their right hand,''} and the simplified version is \emph{``A person sits on a stool, touching the ground with each hand alternately.''}
Next, we employ GPT-4 to classify each description into one of our predefined 15 action labels with in-context learning~\cite{brown2020language}. The action label for the above sequence is \emph{``Sit.''} We meticulously review all generated texts and action labels to ensure high quality and alignment across the dataset.

\noindent\textbf{Unifying Human Representations.} 
Different datasets employ varying human models (\eg, SMPL-H~\cite{MANO}, SMPL-X~\cite{SMPL-X:2019}) and diverse shapes.
A straightforward solution can be to convert different humans from SMPL-H and SMPL-X to a consistent SMPL version and encode shape parameters into the generative modeling. However, although SMPL is widely used in various human-related tasks, it is fundamentally a rotation-based representation. In the context of human motion generation, Cartesian features like joint positions and velocities are more commonly used, as seen in the integration with the HumanML3D representation~\cite{guo2022generating} and in most text-to-motion work~\cite{petrovich22temos, tevet2022human, zhang2023t2m}. 
This is still suboptimal because joints are located beneath the body's surface and do not explicitly participate in interactions.
To overcome the limitations, we use \emph{markers} -- specific sets of human vertices representing human motion and interactions -- as a simple and unified representation capable of effectively inferring contact, evaluated in Table~\ref{tab:smpl_marker_joint}. Similar approaches are discussed in~\cite{wu2022saga, xu2023interdiff}. Then we need to select a marker set that is consistent between SMPL-H and SMPL-X models.

Given two human body models, SMPL-H and SMPL-X, sharing the same shape, we establish marker correspondences in two steps. First, we index the markers on the SMPL-H surface as defined in prior work \cite{wu2022saga, xu2023interdiff}. Second, we locate the corresponding vertices on SMPL-X by selecting the closest points to these SMPL-H markers, leveraging the official SMPL conversion, which maps each SMPL-H vertex to the nearest point on the SMPL-X mesh.
We extensively evaluated the approximation error of these marker correspondences across a broad range of poses. Our results show that the maximum error consistently remains below 1\,cm, a deviation unlikely to affect the overall performance of HOI generation. This high consistency arises because the markers are rigidly attached to the body, and soft deformations are disabled. As a result, the identical rigid transformations in SMPL-H and SMPL-X preserve the correspondence of the markers. Additional details on such correspondence-preserving conditions can be found in \cite{keller2023skin}.
Figure~\ref{fig:marker} illustrates the marker sets for SMPL-H and SMPL-X. We use this marker-based representation to train the generative models for the tasks outlined in Sec.~\ref{sec:tasks}, while still relying on the original SMPL-H or SMPL-X representations for the interaction correction and augmentation methods described in Sec.~\ref{sec:correction}.

\subsection{Interaction Correction and Augmentation}
\label{sec:correction}
In this section, we present a unified optimization framework that addresses both the correction of MoCap artifacts and the augmentation of the dataset by introducing more synthetic data. The process takes as input the motions and geometries of humans and objects, then compares them against predefined standards to define loss functions. Using gradient-based optimization, we iteratively adjust human and object motions to minimize these losses, thereby refining the data to meet the desired quality criteria. The key challenge lies in formulating learning objectives that not only rectify existing data but also facilitate the generation of new synthetic data.

Our optimization is carried out in three sequential steps: (\textbf{1}) full-body correction; (\textbf{2}) hand correction; and (\textbf{3}) interaction augmentation. Hand correction is handled separately because, although hand poses occupy considerable space in the SMPL representation, they contribute relatively little to the overall scale of the learning objectives. By decoupling hand correction from full-body correction, we can better balance these two processes and define more targeted objectives. In what follows, we first introduce the hand correction stage.

\noindent \textbf{Hand Correction.} 
Given that many existing datasets contain inaccurate hand poses~\cite{bhatnagar22behave,li2023object}, our approach selectively promotes contact only in regions where ground-truth data indicates hand-object interaction, while ensuring the hand motion remains natural, in spirit to InterMimic~\cite{xu2025intermimic} but relying on predefined optimization instead of RL.
This approach is effective for the whole-body interaction datasets we utilize, which generally do not require high dexterity and typically only involve the hand conforming to the object for grasping, as a common assumption in existing work~\cite{taheri2024grip, zhou2022toch}, while we distinguish our approach from those that employ multi-stage, learning-based methods for the same purpose.

We divide our hand correction objectives into two categories: \emph{contact promotion} and \emph{hand constraints}. Contact promotion is guided by the following contact loss:
\[
E_{\mathrm{cont}} = \sum_{i=1}^{L} c_i \sum_{j} d_j[i],
\]
where \(d_j[i]\) is the distance between the \(j\)-th hand vertex and its nearest point on the object's surface at the \(i\)-th frame, and \(c_i\) indicates whether the object and the hand are in contact at frame \(i\). The contact indicator \(c_i\), inferred from ground truth data, is a function based on hand-object distance \(\min_j d_j[i]\), which we provide details in supplementary.
The hand constraint objectives are introduced to preserve naturalness and temporal smoothness in the hand motions. These constraints include: (\textbf{1}) penetration loss, which penalizes intersections between the hand and the object.
(\textbf{2}) smoothness loss, which promotes consistent contact and reduces jittering.
(\textbf{3}) prior loss, which constrains the range of motion (RoM) of the fingers to maintain realism. Without this constraint, contact promotion could inadvertently drive fingers into biologically impossible poses.
Detailed formulations of these loss functions are provided in the supplementary.

\noindent \textbf{Full-Body Correction.} In this stage, all human and object poses can be updated via gradient descent. We add a reconstruction loss to ensure that the optimized interactions closely match the ground truth. Other losses mirror those used in hand correction, with two key differences: (\textbf{1}) Contact and penetration losses are computed for the entire body rather than just the hands. (\textbf{2}) Prior loss is omitted because the reconstruction loss alone suffices to maintain plausible human motion. Detailed formulations of these losses are provided in supplementary.

\noindent\textbf{Interaction Augmentation.}
Synthetic data has become increasingly important in computer vision and generative modeling \cite{nguyen2024dataset,chen2024allava,fan2024scaling,black2023bedlam}, prompting a key question in the context of HOI animation: Can we scale up datasets without collecting additional MoCap data? Does existing interaction data offer information beyond its observable motions?
Consider a scenario where a person grasps a box and walks, as illustrated in Figure~\ref{fig:augment}. Even if their gait changes slightly, hand-box contact should remain consistent to preserve the semantics of the interaction. This illustrates the \emph{principle of interaction invariance}: the core interaction persists despite minor variations in motion. Leveraging this principle, we can augment our dataset by injecting new human motions while preserving consistent object interactions. Training neural networks on such augmented data enables them to naturally learn this invariance, a common strategy in symmetric learning \cite{dieleman2016exploiting}, and ultimately enhances model performance.

Our augmentation pipeline consists of three steps:
(\textbf{1}) \textit{Object Displacement}: We apply a random displacement to the object's trajectory, uniformly across all timesteps.
(\textbf{2}) \textit{Interaction Alignment}: We optimize the human motion to maintain interaction with the displaced object, using both contact consistency and reconstruction objectives.  
(\textbf{3}) \textit{Interaction Filtering}: We remove low-quality augmentations, those with unreasonable initial displacements, significant penetrations (human-object or self-penetration), or alignment failures indicated by high optimization losses (\eg, excessive jitter).

During the alignment phase, the primary objective is the \emph{contact consistency loss}. We first compute a distance matrix \(\mathbf{D}\), where each element \(\mathbf{D}_{jk} = \|\mathbf{v}_h^j - \mathbf{v}_o^k\|\) denotes the Euclidean distance between the \(j\)-th human vertex and the \(k\)-th object vertex, before or after displacement. With a reference matrix \(\hat{\mathbf{D}}\) from the original (pre-displacement) setup, we optimize the human motion using:
\[
E_{\mathrm{align}} = \sum_{i=1}^{L} \sum_{j,k} \frac{1}{(\hat{\mathbf{D}}_{jk} + \epsilon)^2} \, \bigl|\hat{\mathbf{D}}_{jk} - \mathbf{D}_{jk}\bigr|^2,
\]
where \(\epsilon\) is a small constant to prevent division by zero. This formulation preserves distances between vertex pairs that were initially close, while de-emphasizing pairs that were farther apart. Additional terms in the objective enforce naturalness and stability in non-interactive regions of the human pose, as detailed in the supplementary material.
\section{Tasks and Methods}
\label{sec:tasks}
In this section, we formally define six distinct tasks featured in our benchmark. We use unified representation across five kinematic generative tasks, where each human-object interaction sequence is represented as \( \langle \boldsymbol{h}, \boldsymbol{o} \rangle \), annotated with an action category \( \boldsymbol{a} \) and a text description \( \boldsymbol{t} \). The human \( \boldsymbol{h} \) includes marker coordinates, marker velocities, signed distance vectors from each marker to the object, and foot-ground contact labels. The object \( \boldsymbol{o} \) represents object motion, including object rotation angles, object translations. Object geometry is described by Basis Point Set (BPS)~\cite{prokudin2019efficient}.

\begin{figure*}
    \centering
    \includegraphics[width=\textwidth]{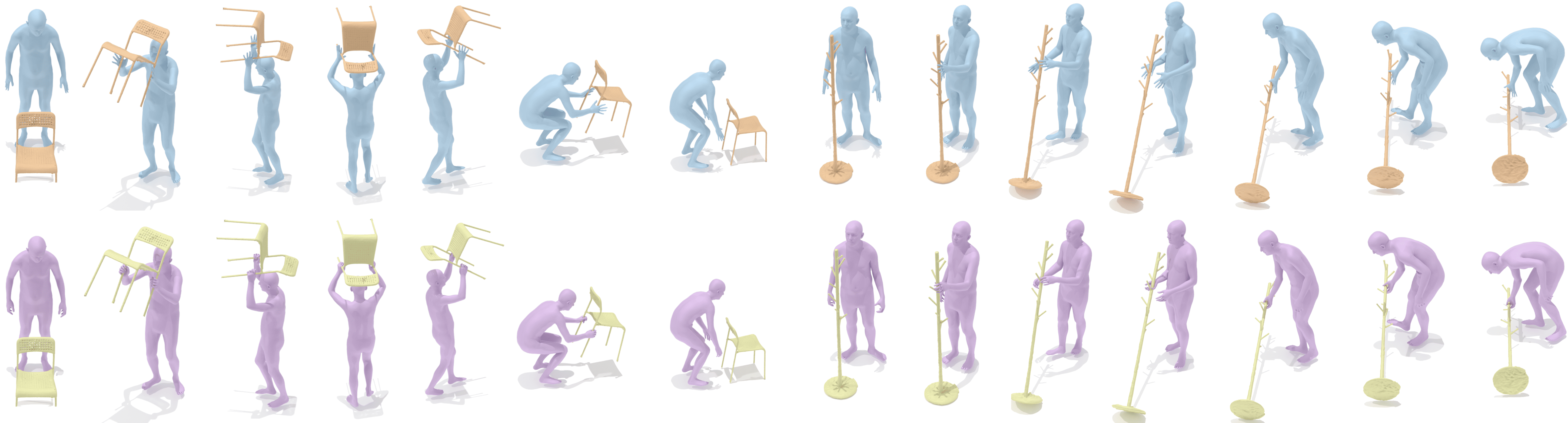}
    \caption{\textbf{Qualitative evaluation} of interaction correction (\textcolor{violet}{bottom}) on the OMOMO~\cite{li2023object} dataset shows hand recovery compared to the ground truth interaction (\textcolor{blue}{top}). Zoom in to see details of the hand recovery.}
    \label{fig:recover}
\end{figure*}

(\textbf{1}) \noindent\textit{Text-Conditioned Interaction Generation.} Initially in~\cite{diller2023cg, peng2023hoi}, the task learns a function to generate the interaction sequence based on text:
$
    \mathcal{G}_{\mathrm{t2i}}(\boldsymbol{t}) \mapsto \langle \boldsymbol{h}, \boldsymbol{o} \rangle
$.

(\textbf{2}) \noindent\textit{Action-Conditioned Interaction Generation.} The objective is to learn a function that maps an action label to the corresponding interaction sequence:
$
    \mathcal{G}_{\mathrm{a2i}}(\boldsymbol{a}) \mapsto \langle \boldsymbol{h}, \boldsymbol{o} \rangle
$.

(\textbf{3}) \noindent\textit{Object-Conditioned Human Generation.} Initially in~\cite{li2023object}, the task generates human motion based on object sequences through a function $
    \mathcal{G}_{\mathrm{o2h}}(\boldsymbol{o}) \mapsto \boldsymbol{h}
$.

(\textbf{4}) \noindent\textit{Human-Conditioned Object Generation.} Conversely, this task focuses on generating object motion sequences from human motion sequences via a function $
    \mathcal{G}_{\mathrm{h2o}}(\boldsymbol{h}) \mapsto \boldsymbol{o}
$.

(\textbf{5}) \noindent\textit{Interaction Prediction.}  
Initially in~\cite{xu2023interdiff}, the task aims to predict future human-object interactions based on past. Let $\langle \boldsymbol{h}_p, \boldsymbol{o}_p \rangle$ denote the past interaction and $\langle \boldsymbol{h}_f, \boldsymbol{o}_f \rangle$ the future. The goal is to learn:
$\mathcal{G}_{\mathrm{p2f}}(\langle \boldsymbol{h}_p, \boldsymbol{o}_p \rangle) \mapsto \langle \boldsymbol{h}_f, \boldsymbol{o}_f \rangle
$.

(\textbf{6}) \noindent\textit{Interaction Imitation.}  
Following~\cite{wang2023physhoi,zhang2023simulation,bae2023pmp}, this task focuses on learning physics-based control policies to reproduce human-object interactions in a physics simulator.
The output is an action sequence $\boldsymbol{f}$, specified as joint Proportional-Derivative (PD) targets. The goal is to learn a function $\mathcal{G}_{\mathrm{i2f}}$ that maps the reference interaction sequences to the PD actuation sequences:
$\mathcal{G}_{\mathrm{i2f}}(\langle \boldsymbol{h}, \boldsymbol{o} \rangle) \mapsto \boldsymbol{f}
$.

\noindent\textbf{Unifying Multi-Task HOI Generation.}
We introduce an additional feature, \(\boldsymbol{\eta}\), which encodes human-object relationships through vectors extending from each human marker to its nearest point on the object's surface. The specific configuration of \(\boldsymbol{\eta}\) for each generative task is described in Sec.~\ref{sec:experiments}.
Using this feature, we can unify first five kinematic generative tasks into a multi-task learning framework by treating \(\boldsymbol{\eta}\) as an additional output. For example, we redefine the text-conditioned interaction task as \(\mathcal{G}_{\mathrm{t2i}}(\boldsymbol{t}) \mapsto \langle \boldsymbol{h}, \boldsymbol{o}, \boldsymbol{\eta}\rangle\), where \(\mathcal{G}\) is a transformer-based diffusion model. This formulation compels the model to learn spatial relationships inherent to the interactions. In our experiments, we observe that this simple strategy, enhanced by large-scale data, consistently outperforms existing methods. Similar ideas are explored in \cite{diller2023cg,li2023controllable,wu2024thor,song2024hoianimator,peng2023hoi,xu2023interdiff,li2023object}.

\section{Experiments} \label{sec:experiments}
We begin by evaluating the effectiveness of our data correction and augmentation methods. Following this, we benchmark existing work and our proposed method on the tasks using our dataset. We standardize the evaluation metrics and present extensive results, including ablation studies. We include additional implementation details, such as the train-test split, in supplementary.

\subsection{Correction and Augmentation}

\noindent\textbf{Metrics.} We use the following two metrics: \textbf{Penetration} refers to the intersection depth -- maximum of negative sign distances from human vertices to the object’s surface -- average across the sequence. \textbf{Contact Ratio} represents the average ratio of human vertices where their distances to object are under a threshold.
\begin{table}
    \centering
    \resizebox{\columnwidth}{!}{
    \begin{tabular}{@{}lcccccccc@{}}
        \toprule
        {Dataset} & {Correction } & Augmentation &   {Pene} (m)$^\downarrow$  & {Cont Ratio} & {User Study} 
         ($\%$)\\ 

        \midrule

        \multirow{3}{*}{BEHAVE~\cite{bhatnagar22behave}} & $\times$ & $\times$& 0.017 & 0.048 & 22.3  \\
        & $\checkmark$ &$\times$& \textbf{0.016} & {0.071} & \textbf{39.7}   \\
        & $\checkmark$ &$\checkmark$& \textbf{0.016} & {0.069} &  38.0 \\

        \midrule

         \multirow{3}{*}{OMOMO~\cite{li2023object}} & $\times$ &$\times$& 0.009 & 0.071 &23.9  \\

         & $\checkmark$ &$\times$& \textbf{0.007} & 0.131  & \textbf{39.4}  \\
         
         & $\checkmark$ & $\checkmark $ & {0.011} & {0.137} &36.7\\

        \bottomrule
    \end{tabular}}
    \caption{\textbf{Quantitative evaluation and user study} on the quality of data from interaction correction and augmentation. }
    \label{tab:corr_aug_quantitative}
    \vspace{-1em}
\end{table}
\begin{figure*}
    \centering
    \includegraphics[width=\textwidth]{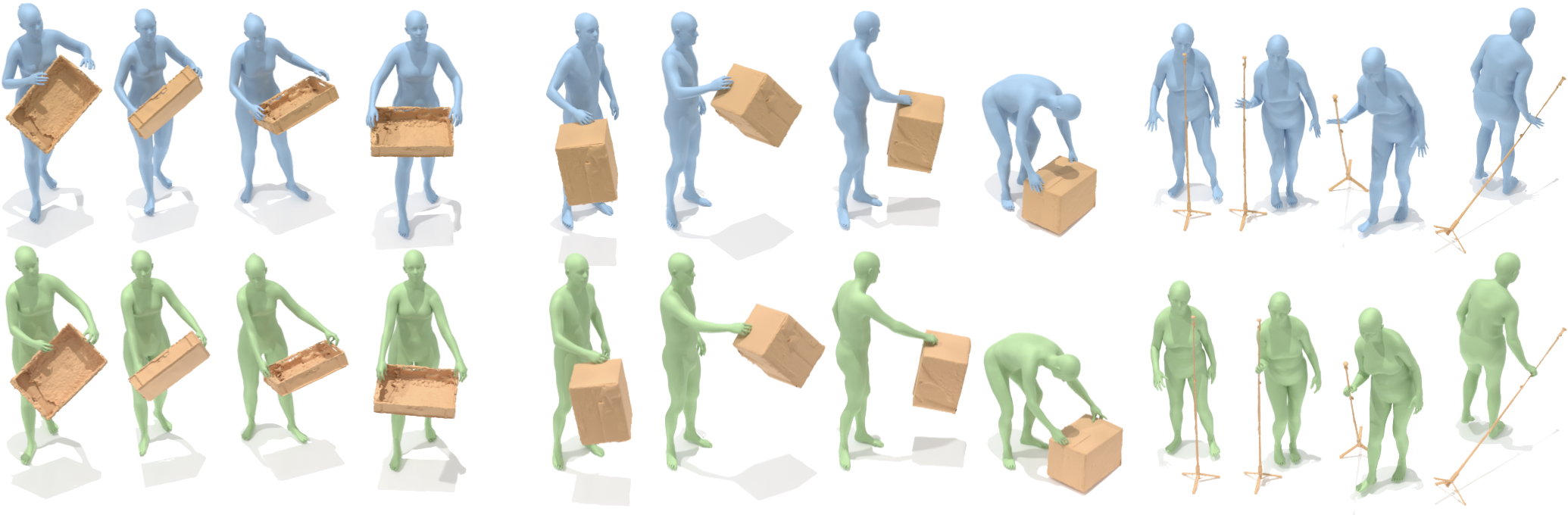}
    \caption{\textbf{Qualitative evaluation} of interaction augmentation (\textcolor{green}{bottom}) shows high-quality synthetic data varied from original (\textcolor{blue}{top}).}
    \label{fig:augment}
\end{figure*}
\begin{table*}
    \centering

    \resizebox{\textwidth}{!}{
    \begin{tabular}{@{}ccccccccccc@{}}
        \toprule
         \multirow{2}{*}{\makecell{HOI-Aware \\ Object Enc.}} & \multirow{2}{*}{\makecell{HOI-Aware \\ Text Enc.}} &\multirow{2}{*}{\makecell{Contact \\ Generation}} & \multirow{2}{*}{\makecell{Contact \\ Guidance}} &\multicolumn{3}{c}{R-Precision$^\uparrow$} & \multirow{2}{*}{FID$^\downarrow$} & \multirow{2}{*}{MM Dist$^\downarrow$} & \multirow{2}{*}{Multimodality$^\uparrow$} & \multirow{2}{*}{Diversity$^\rightarrow$} \\
        
        \cmidrule(lr){5-7} 
        &  &  & & Top 1 & Top 2 & Top 3 & & &  \\
        \midrule
        \multicolumn{4}{c}{Ground Truth} &  0.852$^{\pm0.000}$ & 0.966$^{\pm0.001}$& 0.989$^{\pm0.001}$ & 0.000$^{\pm0.000}$ & 2.810$^{\pm0.002}$  & - &11.489$^{\pm0.011}$  \\
        \midrule
        \xmark & \xmark & \xmark & \xmark &   0.733$^{\pm0.007}$ & 0.909$^{\pm0.002}$ & 0.957$^{\pm0.002}$ & 3.192$^{\pm0.191}$ & 4.950$^{\pm0.023}$ & 3.149$^{\pm0.452}$ &  11.192$^{\pm0.019}$  \\
        \xmark & \xmark & \cmark & \xmark &  0.730$^{\pm0.007}$ &  0.913$^{\pm0.004}$ & 0.958$^{\pm0.005}$ & 1.997$^{\pm0.092}$ & 4.752$^{\pm0.065}$ & \textbf{4.171}$^{\pm0.027}$ & \textbf{11.501}$^{\pm0.037}$   \\
        \cmark & \xmark & \cmark & \xmark & 0.737$^{\pm0.011}$ & 0.912$^{\pm0.002}$ & 0.963$^{\pm0.008}$ & 1.837$^{\pm0.126}$ & 4.631$^{\pm0.078}$ & 2.836$^{\pm0.583}$ & 11.369$^{\pm0.096}$  
        \\
        \cmark & \cmark & \cmark & \xmark & \textbf{0.784}$^{\pm0.004}$ & \textbf{0.940}$^{\pm0.002}$ & \textbf{0.980}$^{\pm0.003}$ & 1.570$^{\pm0.139}$ & 4.414$^{\pm0.064}$ & 2.677$^{\pm0.562}$ & 11.409$^{\pm0.005}$  \\   
        \cmark & \cmark & \cmark & \cmark & \textbf{0.784}$^{\pm0.004}$ & \textbf{0.940}$^{\pm0.000}$ & 0.977$^{\pm0.002}$ & \textbf{1.567}$^{\pm0.144}$ & \textbf{4.412}$^{\pm0.065}$ & 3.842$^{\pm0.005}$ & 11.518$^{\pm0.178}$  \\
        \bottomrule     

    \end{tabular}}
    \caption{\textbf{Quantitative evaluation} on the task of text-conditioned interaction generation. A batch size of 64 is used for R-Precision.}
    \label{tab:t2i_experiment}
    \vspace{-0.5em}
\end{table*}

\noindent\textbf{Quantitative Evaluations.} Table~\ref{tab:corr_aug_quantitative} shows that our correction process significantly improves the quality of the original MoCap data by enhancing human-object contact and reducing penetration artifacts. Moreover, the quality of the augmented data is comparable to that of the corrected data and exceeds the quality of the original dataset.

\noindent\textbf{Qualitative Evaluations.}
Recognizing that quantitative metrics may not fully capture data quality, we conducted a \emph{double-blind} user study. We randomly selected sequences from each of the raw, corrected, and augmented data for the subset from BEHAVE~\cite{bhatnagar22behave} and OMOMO~\cite{li2023object} datasets. Human judges were presented with 30 tuple of interactions and asked to rank the quality of three sequences. According to Table~\ref{tab:corr_aug_quantitative}, over 39\% of judges select the corrected data as having the highest quality, significantly outperforming the original data. This confirms that our correction process effectively enhances data realism. Moreover, the augmented data receive ratings comparable to the corrected data, indicating that our synthetic data is of high quality.
In Figure~\ref{fig:recover}, we visualize our correction results. Despite the original data lacking detailed hand information, we successfully recover vivid and accurate hand interactions. Figure~\ref{fig:augment} showcases our augmentation, which introduces new high-quality synthetic data while maintaining consistent contact in interactions.

\subsection{Language Conditioned HOI Generation}  \label{sec:language}

\noindent \textbf{Metrics.}
Following the literature on text-to-motion generation~\cite{guo2022generating}, we develop five metrics for evaluation. The Fréchet Inception Distance (\textbf{FID}) quantifies the similarity between generated HOI features and the ground truth. The \textbf{Multimodality} and \textbf{Diversity} metrics assess the variety within the generated HOI. \textbf{R-Precision} measures the alignment between the textual descriptions and the generated HOI. The Multimodal Distance (\textbf{MM Dist}) evaluates the disparity between HOI features and corresponding text features.
To obtain human-object interaction (HOI) and text features for calculating these metrics, existing methods often train their feature extractors on very limited data~\cite{peng2023hoi,diller2023cg,wu2024thor,song2024hoianimator}, which can degrade the quality of the evaluation. We address this limitation by incorporating our larger-scale data with marker and BPS representations. Instead of formulating a classification task to train the feature extractor~\cite{guo2022generating}, we follow~\cite{humantomato,petrovich2023tmr} and employ sequence-level contrastive learning with an InfoNCE loss~\cite{oord2018representation} to train a text encoder and an HOI encoder, integrating Sentence-BERT~\cite{reimers2019sentence} into the text encoder. 

\noindent \textbf{Baselines and Implementation Details.}
We adopt HOI-Diff~\cite{peng2023hoi} as our base model because it is the only publicly available option compatible with our requirements. For example, CHOIS~\cite{li2023controllable} requires additional conditions beyond text input. HOI-Diff utilizes a transformer-based diffusion model~\cite{tevet2022human} as the backbone, and integrates an affordable model as the classifier guidance~\cite{dhariwal2021diffusion}. We develop several baseline variants towards our final method by implementing three key modifications:
(\textbf{i}) Text Encoder: We replace HOI-Diff's CLIP-based text encoder, where the latent space is not structured for human-object interactions, with our pretrained interaction-aware text encoder.
(\textbf{ii}) Object Shape Encoding: We substitute the original PointNet++~\cite{qi2017pointnet++}, also not pretrained for capturing interaction, with BPS~\cite{prokudin2019efficient} representations for encoding object shapes.
(\textbf{iii}) Instead of using affordance as guidance, we regress the contact representation \(\boldsymbol{\eta}\) through a multi-task learning. (\textbf{iv}) We further incorporate the contact prediction as classifier guidance~\cite{dhariwal2021diffusion} within the denoising process, which we detail in supplementary.
\begin{table}
    \centering

    \resizebox{0.8\columnwidth}{!}{
    \begin{tabular}{@{}cccc@{}}
        \toprule
        Method & FID$^\downarrow$  & Multimodality$^\uparrow$ & Diversity$^\rightarrow$ \\
        \midrule
        
        Ground Truth & 
        0.000$^{\pm0.000}$ & 
        - & 
        11.489$^{\pm0.011}$ \\
        \midrule
        HOI-Diff~\cite{peng2023hoi} &
        3.566$^{\pm0.098}$ & 
        5.321$^{\pm0.143}$ & 
        10.989$^{\pm0.112}$ \\
        \textbf{Ours} & \textbf{2.161}$^{\pm0.037}$ & \textbf{5.792}$^{\pm0.059}$ & \textbf{11.291}$^{\pm0.261}$   \\
        \bottomrule     

    \end{tabular}}
    
    \caption{\textbf{Quantitative evaluation} on the task of action-conditioned interaction generation on the entire InterAct testset.}
    \label{tab:a2i_experiment}
    \vspace{-1em}
\end{table}

\noindent \textbf{Quantitative Results.} Table~\ref{tab:t2i_experiment} presents the evaluation results for the text-conditioned interaction generation task. We assess the impact of four design choices introduced above. Notably, incorporating contact modeling and BPS encoding significantly improves the quality of generated HOI, substantially enhancing the FID score. Furthermore, using our interaction-aware text and HOI encoder enhances the quality of the generated interactions and improves the alignment between the generated results and the text input. Lastly, classifier guidance provides a slight overall performance improvement.
Table~\ref{tab:a2i_experiment} benchmarks the performance improvements of these design choices similarly for the action-conditioned interaction generation task.

\noindent \textbf{Effectiveness of Marker-Based Representation.}
As shown in Table~\ref{tab:smpl_marker_joint}, we compare different human representations on text-to-interaction generation. Without altering contact modeling, marker-based representation produces interactions with fewer artifacts compared to other representations.
\begin{table}
    \centering
    \resizebox{0.6\columnwidth}{!}{
    \begin{tabular}{@{}lccc@{}}
        \toprule
        {Representation}  &   {Pene} (m)$^\downarrow$  & {Cont Ratio} \\
        
        \midrule

         SMPL &{0.030} & {0.025}  \\

        \midrule

          Joint & {0.027} & {0.032} \\
          \midrule
          
          Marker & \textbf{0.025} & {0.028} \\

        \bottomrule
    \end{tabular}}
        \caption{\textbf{Ablation study} on different representation for text-to-interaction task, evaluated under BEHAVE~\cite{bhatnagar22behave} subset.   }
        \label{tab:smpl_marker_joint}
\end{table}

\begin{table}[t]
    \begin{minipage}{\linewidth}
    \centering

    \resizebox{\textwidth}{!}{
    \begin{tabular}{@{}ccccccc@{}}
        \toprule
        Model  & MPMPE $^\downarrow$ & FS$^\downarrow$ & $C_\mathrm{prec}$ $^\uparrow$  & $C_\mathrm{rec}$ $^\uparrow$ & $C_\mathrm{acc}$ $^\uparrow$  & F1 Score $^\uparrow$ \\

        \midrule
        
        2-stage~\cite{li2023object}   & 36.50 & \textbf{0.27} & 0.81 & 0.85 & 0.77 & 0.80  \\
        1-stage   & 36.94 & 0.29 & 0.84 & 0.82 & 0.80 & 0.81  \\
        \textbf{Ours} (Disc.)   & 36.95 & 0.28 & \textbf{0.85} & 0.84 & 0.83 & 0.82 \\
        \textbf{Ours} (Cont.)  & \textbf{35.69} & 0.28 & \textbf{0.85} & \textbf{0.89} & \textbf{0.85} & \textbf{0.85} \\

        \bottomrule
    \end{tabular}}

    \end{minipage}
    \hfill
    \caption{\textbf{Quantitative evaluation} on object-conditioned human motion generation, with novel objects unseen from training.}
    \label{tab:o2h_after}
    \vspace{-1em}
\end{table}

\label{sec:hoi_inpaint}
\subsection{HOI Inpainting}

\noindent \textbf{Metrics.} Following OMOMO~\cite{li2023object}, we develop six metrics tailored for evaluating our marker-based representation. The Mean Per-Marker Position Error (\textbf{MPMPE}) are used to measure the similarity between the generated marker motion and the ground truth. Foot Sliding (\textbf{FS}) is employed to assess the skating effect, reflecting the plausibility of the motion. Additionally, we use a set of contact metrics, including precision ($C_\mathrm{prec}$), recall ($C_\mathrm{rec}$), accuracy ($C_\mathrm{acc}$), and the F1 Score, to evaluate the quality of human-object contact compared to the ground truth. For the human-conditioned object generation task, we include $T_{\mathrm{err}}$ and $O_{\mathrm{err}}$, which measure discrepancies in object translation and orientation between the generated and the ground truth.

\noindent \textbf{Baselines and Implementation Details.} We adopt OMOMO~\cite{li2023object} as our base model since it is the only publicly available option. OMOMO employs a two-stage generation process: it first generates hand motions and then uses these to guide full-body generations. This strategy is particularly effective for the OMOMO dataset, where interactions primarily involve hand contact. However, it is less effective when applied to our InterAct data, which features more versatile whole-body engagement. Motivated by this, we use a single-stage pipeline with multi-task learning, as introduced in Sec.~\ref{sec:tasks}.
We investigate two different choices for contact regression:
$\boldsymbol{\eta}_{\mathrm{Cont.}}$, which encodes the nearest vector and distance between human markers and the object.
$\boldsymbol{\eta}_{\mathrm{Disc.}}$, which encodes the contact labels for each marker.

\noindent \textbf{Quantitative Results.} Tables~\ref{tab:o2h_after} and~\ref{tab:h2o_after} illustrate the effectiveness of our single-stage pipeline, with notable performance improvements when incorporating multi-task modeling. These results provide additional evidence, complementing the evaluation of text-conditioned generation, that multi-task learning significantly enhances model performance.

\subsection{Interaction Prediction}
\noindent \textbf{Evaluation Metrics.} We compare the generated poses to the ground truth motion data using \textbf{MPMPE} (mean per-marker errors), measured in meters. Second, we assess object motion accuracy using \textbf{Trans.~Err.}, the average $l_2$ distance between the predicted and ground truth object translations, and \textbf{Rot.Err.}, the average $l_1$ distance between the predicted and ground truth object quaternions, following~\cite{xu2023interdiff}.

\noindent \textbf{Implementation Details.} We adapt InterDiff~\cite{xu2023interdiff} to use a marker-based representation and evaluate it at multiple transformer latent dimensions (512, 1024, and 1536), keeping all other components consistent with the original implementation. This allows us to assess that model trained our data benefits from scaling laws.

\subsection{Interaction Imitation}
\begin{figure}
    \centering
    \includegraphics[width=\columnwidth]{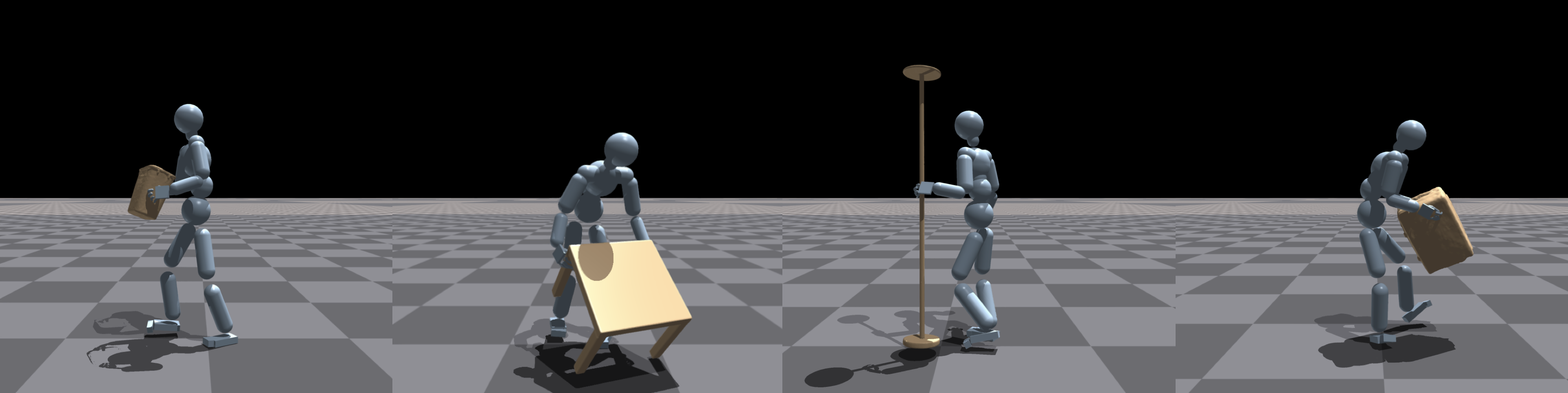}
    \caption{Qualitative results demonstrate the successful imitation of our corrected data using PhysHOI~\cite{wang2023physhoi}.
    }
    \label{fig:imitation}
\end{figure}
\noindent\textbf{Implementation Details.} We use PhysHOI~\cite{wang2023physhoi} to imitate sequences from our InterAct dataset, selecting four sequences shown in Figure~\ref{fig:imitation}. The IsaacGym~\cite{makoviychuk2021isaac} simulator is used, following the same architecture, reward, and representation design as PhysHOI. Training for each sequence, with separate evaluations for both raw and corrected data, is performed on a single NVIDIA A40 GPU over the course of one day.

\noindent\textbf{Quantitative Evaluation.} In addition to Figure~\ref{fig:imitation}, which presents the qualitative results, we evaluate the imitation policy by training on both corrected and raw data, reporting the success rate as defined in~\cite{wang2023physhoi}. Evaluating four examples from Figure~\ref{fig:imitation} and averaging the success rates over 2048 environments, training on the corrected data achieves a success rate of 90.7\%, surpassing the 84.4\% achieved with raw data. This demonstrates that our interaction correction provides better data for the motion imitation task.

\noindent \textbf{Quantitative Results.} Table~\ref{tab:p2f_experiment} demonstrates that our dataset, with its larger volume of data, supports the improved performance of the trained model with larger scale. In contrast, training the model on limited data leads to overfitting.

\begin{table}[t]
    \begin{minipage}{\linewidth}
    \centering
    \resizebox{\textwidth}{!}{
    \begin{tabular}{@{}cccccccc@{}}
        \toprule
        Model   & $T_{\mathrm{err}}$ $^\downarrow$ & $O_{\mathrm{err}}$ $^\downarrow$ & $C_\mathrm{prec}$ $^\uparrow$  & $C_\mathrm{rec}$ $^\uparrow$ & $C_\mathrm{acc}$ $^\uparrow$  & F1 Score $^\uparrow$ \\

        \midrule
        
         1-stage  & 25.92 &	0.91 &	0.83 &	0.66 &	0.72 &	0.69   \\
        \textbf{Ours} (multi-task)  &	\textbf{23.98} &	\textbf{0.83} &	\textbf{0.84} &	\textbf{0.68} &	\textbf{0.74} &	\textbf{0.72}   \\

        \bottomrule
    \end{tabular}}

    \end{minipage}
    \hfill
    \caption{\textbf{Quantitative evaluation} on human-to-object. }
    \label{tab:h2o_after}
\end{table}
\begin{table}
    \centering

    \resizebox{\columnwidth}{!}{
    \begin{tabular}{@{}ccccccc@{}}
        \toprule
        Training data & Model size & Global MPMPE$^\downarrow$ & Local MPMPE$^\downarrow$ & Trans. Err.$^\downarrow$  & Rot.Err.$^\downarrow$ \\
        \midrule
        \multirow{3}{*}{BEHAVE}&$\times1$ & 0.120 & 0.103 & 0.133 & 0.352\\
        \cmidrule{2-6}
        &$\times2$ & 0.105 & 0.092 & 0.109 & 0.312 \\
        \cmidrule{2-6}
        &$\times3$ & 0.113 & 0.100 & 0.118 & 0.343 \\
         \midrule
        \multirow{3}{*}{InterAct (\textbf{Ours})}& $\times1$& 0.106 &  0.095 & 0.106 & 0.297 \\
         \cmidrule{2-6}
         & $\times2$ & 0.094 & 0.083 & 0.103 & 0.286 \\
      \cmidrule{2-6}
        &$\times3$ & \textbf{0.091} & \textbf{0.079} & \textbf{0.094} & \textbf{0.264}\\

        \bottomrule     
    \end{tabular}}
    \caption{\textbf{Quantitative evaluation} on interaction prediction.}
    \label{tab:p2f_experiment}
\end{table}

\section{Conclusion} \label{sec:conclusion}
We introduce InterAct, a large-scale 3D whole-body human-object interaction benchmark. We employ a unified optimization framework that performs interaction correction and augmentation, enhancing data quality and augment the dataset with synthetic data, scaling to 30.70 hours of interactions and 48,630 textual descriptions. We introduce a simple yet effective multi-task learning approach for unified HOI modeling, enabling models to be trained more effectively across multiple tasks. Our comprehensive experiments highlight the significant advantages of InterAct and our methodologies, resulting in more expressive interaction generation.
\newpage
\paragraph{Acknowledgments.} This work was supported in part by NSF Grant 2106825, NIFA Award 2020-67021-32799, the Amazon-Illinois Center on AI for Interactive Conversational Experiences, the Toyota Research Institute, the IBM-Illinois Discovery Accelerator Institute, and Snap Inc. This work used computational resources, including the NCSA Delta and DeltaAI and the PTI Jetstream2 supercomputers through allocations CIS230012, CIS230013, and CIS240311 from the Advanced Cyberinfrastructure Coordination Ecosystem: Services \& Support (ACCESS) program, as well as the TACC Frontera supercomputer, Amazon Web Services (AWS), and OpenAI API through the National Artificial Intelligence Research Resource (NAIRR) Pilot.

{
    \small
    \bibliographystyle{ieeenat_fullname}
    \bibliography{main}

\begin{thebibliography}{127}
\providecommand{\natexlab}[1]{#1}
\providecommand{\url}[1]{\texttt{#1}}
\expandafter\ifx\csname urlstyle\endcsname\relax
  \providecommand{\doi}[1]{doi: #1}\else
  \providecommand{\doi}{doi: \begingroup \urlstyle{rm}\Url}\fi

\bibitem[CMU()]{CMU-Mocap}
{CMU} graphics lab motion capture database.
\newblock {\url{http://mocap.cs.cmu.edu/}}.

\bibitem[eas(2021)]{easymocap}
Easymocap - make human motion capture easier.
\newblock Github, 2021.

\bibitem[Bae et~al.(2023)Bae, Won, Lim, Min, and Kim]{bae2023pmp}
Jinseok Bae, Jungdam Won, Donggeun Lim, Cheol-Hui Min, and Young~Min Kim.
\newblock Pmp: Learning to physically interact with environments using part-wise motion priors.
\newblock In \emph{SIGGRAPH}, 2023.

\bibitem[Bhatnagar et~al.(2022)Bhatnagar, Xie, Petrov, Sminchisescu, Theobalt, and Pons-Moll]{bhatnagar22behave}
Bharat~Lal Bhatnagar, Xianghui Xie, Ilya Petrov, Cristian Sminchisescu, Christian Theobalt, and Gerard Pons-Moll.
\newblock {BEHAVE}: Dataset and method for tracking human object interactions.
\newblock In \emph{CVPR}, 2022.

\bibitem[Black et~al.(2023)Black, Patel, Tesch, and Yang]{black2023bedlam}
Michael~J Black, Priyanka Patel, Joachim Tesch, and Jinlong Yang.
\newblock Bedlam: A synthetic dataset of bodies exhibiting detailed lifelike animated motion.
\newblock In \emph{CVPR}, 2023.

\bibitem[Braun et~al.(2023)Braun, Christen, Kocabas, Aksan, and Hilliges]{braun2023physically}
Jona Braun, Sammy Christen, Muhammed Kocabas, Emre Aksan, and Otmar Hilliges.
\newblock Physically plausible full-body hand-object interaction synthesis.
\newblock \emph{arXiv preprint arXiv:2309.07907}, 2023.

\bibitem[Brown et~al.(2020)Brown, Mann, Ryder, Subbiah, Kaplan, Dhariwal, Neelakantan, Shyam, Sastry, Askell, et~al.]{brown2020language}
Tom Brown, Benjamin Mann, Nick Ryder, Melanie Subbiah, Jared~D Kaplan, Prafulla Dhariwal, Arvind Neelakantan, Pranav Shyam, Girish Sastry, Amanda Askell, et~al.
\newblock Language models are few-shot learners.
\newblock In \emph{NeurIPS}, 2020.

\bibitem[Cao et~al.(2020)Cao, Gao, Mangalam, Cai, Vo, and Malik]{cao2020long}
Zhe Cao, Hang Gao, Karttikeya Mangalam, Qi-Zhi Cai, Minh Vo, and Jitendra Malik.
\newblock Long-term human motion prediction with scene context.
\newblock In \emph{ECCV}, 2020.

\bibitem[Cha et~al.(2024)Cha, Kim, Yoon, and Baek]{cha2024text2hoi}
Junuk Cha, Jihyeon Kim, Jae~Shin Yoon, and Seungryul Baek.
\newblock Text2hoi: Text-guided 3d motion generation for hand-object interaction.
\newblock In \emph{CVPR}, 2024.

\bibitem[Chao et~al.(2021{\natexlab{a}})Chao, Yang, Chen, and Deng]{chao2021learning}
Yu-Wei Chao, Jimei Yang, Weifeng Chen, and Jia Deng.
\newblock Learning to sit: Synthesizing human-chair interactions via hierarchical control.
\newblock In \emph{AAAI}, 2021{\natexlab{a}}.

\bibitem[Chao et~al.(2021{\natexlab{b}})Chao, Yang, Xiang, Molchanov, Handa, Tremblay, Narang, Van~Wyk, Iqbal, Birchfield, et~al.]{chao2021dexycb}
Yu-Wei Chao, Wei Yang, Yu Xiang, Pavlo Molchanov, Ankur Handa, Jonathan Tremblay, Yashraj~S Narang, Karl Van~Wyk, Umar Iqbal, Stan Birchfield, et~al.
\newblock Dexycb: A benchmark for capturing hand grasping of objects.
\newblock In \emph{CVPR}, 2021{\natexlab{b}}.

\bibitem[Chen et~al.(2024)Chen, Chen, Zhang, Chen, Wu, Zhang, Chen, Li, Wan, and Wang]{chen2024allava}
Guiming~Hardy Chen, Shunian Chen, Ruifei Zhang, Junying Chen, Xiangbo Wu, Zhiyi Zhang, Zhihong Chen, Jianquan Li, Xiang Wan, and Benyou Wang.
\newblock Allava: Harnessing gpt4v-synthesized data for a lite vision-language model.
\newblock \emph{arXiv preprint arXiv:2402.11684}, 2024.

\bibitem[Christen et~al.(2024)Christen, Hampali, Sener, Remelli, Hodan, Sauser, Ma, and Tekin]{christen2024diffh2o}
Sammy Christen, Shreyas Hampali, Fadime Sener, Edoardo Remelli, Tomas Hodan, Eric Sauser, Shugao Ma, and Bugra Tekin.
\newblock Diffh2o: Diffusion-based synthesis of hand-object interactions from textual descriptions.
\newblock \emph{arXiv preprint arXiv:2403.17827}, 2024.

\bibitem[Corona et~al.(2020)Corona, Pumarola, Alenya, and Moreno-Noguer]{corona2020context}
Enric Corona, Albert Pumarola, Guillem Alenya, and Francesc Moreno-Noguer.
\newblock Context-aware human motion prediction.
\newblock In \emph{CVPR}, 2020.

\bibitem[Cui et~al.(2024)Cui, Liu, Liu, Yang, Zhu, and Huang]{cui2024anyskill}
Jieming Cui, Tengyu Liu, Nian Liu, Yaodong Yang, Yixin Zhu, and Siyuan Huang.
\newblock {AnySkill}: Learning open-vocabulary physical skill for interactive agents.
\newblock In \emph{CVPR}, 2024.

\bibitem[Dhariwal and Nichol(2021)]{dhariwal2021diffusion}
Prafulla Dhariwal and Alexander Nichol.
\newblock Diffusion models beat gans on image synthesis.
\newblock In \emph{NeurIPS}, 2021.

\bibitem[Dieleman et~al.(2016)Dieleman, De~Fauw, and Kavukcuoglu]{dieleman2016exploiting}
Sander Dieleman, Jeffrey De~Fauw, and Koray Kavukcuoglu.
\newblock Exploiting cyclic symmetry in convolutional neural networks.
\newblock In \emph{ICML}, 2016.

\bibitem[Diller and Dai(2024)]{diller2023cg}
Christian Diller and Angela Dai.
\newblock {CG-HOI}: Contact-guided 3d human-object interaction generation.
\newblock In \emph{CVPR}, 2024.

\bibitem[Fan et~al.(2024)Fan, Chen, Krishnan, Katabi, Isola, and Tian]{fan2024scaling}
Lijie Fan, Kaifeng Chen, Dilip Krishnan, Dina Katabi, Phillip Isola, and Yonglong Tian.
\newblock Scaling laws of synthetic images for model training... for now.
\newblock In \emph{CVPR}, 2024.

\bibitem[Fan et~al.(2023)Fan, Taheri, Tzionas, Kocabas, Kaufmann, Black, and Hilliges]{fan2023arctic}
Zicong Fan, Omid Taheri, Dimitrios Tzionas, Muhammed Kocabas, Manuel Kaufmann, Michael~J. Black, and Otmar Hilliges.
\newblock {ARCTIC}: A dataset for dexterous bimanual hand-object manipulation.
\newblock In \emph{CVPR}, 2023.

\bibitem[Ghosh et~al.(2022)Ghosh, Dabral, Golyanik, Theobalt, and Slusallek]{ghosh2022imos}
Anindita Ghosh, Rishabh Dabral, Vladislav Golyanik, Christian Theobalt, and Philipp Slusallek.
\newblock {IMoS}: Intent-driven full-body motion synthesis for human-object interactions.
\newblock \emph{arXiv preprint arXiv:2212.07555}, 2022.

\bibitem[Guo et~al.(2022)Guo, Zou, Zuo, Wang, Ji, Li, and Cheng]{guo2022generating}
Chuan Guo, Shihao Zou, Xinxin Zuo, Sen Wang, Wei Ji, Xingyu Li, and Li Cheng.
\newblock Generating diverse and natural 3d human motions from text.
\newblock In \emph{CVPR}, 2022.

\bibitem[Hampali et~al.(2020)Hampali, Rad, Oberweger, and Lepetit]{hampali2020honnotate}
Shreyas Hampali, Mahdi Rad, Markus Oberweger, and Vincent Lepetit.
\newblock Honnotate: A method for 3d annotation of hand and object poses.
\newblock In \emph{CVPR}, 2020.

\bibitem[Hassan et~al.(2019)Hassan, Choutas, Tzionas, and Black]{hassan2019resolving}
Mohamed Hassan, Vasileios Choutas, Dimitrios Tzionas, and Michael~J Black.
\newblock Resolving 3d human pose ambiguities with 3d scene constraints.
\newblock In \emph{ICCV}, 2019.

\bibitem[Hassan et~al.(2021)Hassan, Ceylan, Villegas, Saito, Yang, Zhou, and Black]{hassan_samp_2021}
Mohamed Hassan, Duygu Ceylan, Ruben Villegas, Jun Saito, Jimei Yang, Yi Zhou, and Michael Black.
\newblock Stochastic scene-aware motion prediction.
\newblock In \emph{ICCV}, 2021.

\bibitem[Hassan et~al.(2023)Hassan, Guo, Wang, Black, Fidler, and Peng]{hassan2023synthesizing}
Mohamed Hassan, Yunrong Guo, Tingwu Wang, Michael Black, Sanja Fidler, and Xue~Bin Peng.
\newblock Synthesizing physical character-scene interactions.
\newblock In \emph{SIGGRAPH}, 2023.

\bibitem[Hou et~al.(2023)Hou, Yu, and Tao]{hou2023compositional}
Zhi Hou, Baosheng Yu, and Dacheng Tao.
\newblock Compositional 3d human-object neural animation.
\newblock \emph{arXiv preprint arXiv:2304.14070}, 2023.

\bibitem[Huang et~al.(2022)Huang, Taheri, Black, and Tzionas]{huang2022intercap}
Yinghao Huang, Omid Taheri, Michael~J. Black, and Dimitrios Tzionas.
\newblock {InterCap}: {J}oint markerless {3D} tracking of humans and objects in interaction.
\newblock In \emph{GCPR}, 2022.

\bibitem[Jiang et~al.(2023)Jiang, Liu, Cao, Cui, Chen, Wang, Zhu, and Huang]{jiang2022chairs}
Nan Jiang, Tengyu Liu, Zhexuan Cao, Jieming Cui, Yixin Chen, He Wang, Yixin Zhu, and Siyuan Huang.
\newblock {CHAIRS}: Towards full-body articulated human-object interaction.
\newblock In \emph{ICCV}, 2023.

\bibitem[Jiang et~al.(2024)Jiang, Zhang, Li, Ma, Wang, Chen, Liu, Zhu, and Huang]{jiang2024scaling}
Nan Jiang, Zhiyuan Zhang, Hongjie Li, Xiaoxuan Ma, Zan Wang, Yixin Chen, Tengyu Liu, Yixin Zhu, and Siyuan Huang.
\newblock Scaling up dynamic human-scene interaction modeling.
\newblock In \emph{CVPR}, 2024.

\bibitem[Keller et~al.(2023)Keller, Werling, Shin, Delp, Pujades, Liu, and Black]{keller2023skin}
Marilyn Keller, Keenon Werling, Soyong Shin, Scott Delp, Sergi Pujades, C~Karen Liu, and Michael~J Black.
\newblock From skin to skeleton: Towards biomechanically accurate 3d digital humans.
\newblock \emph{ACM Transactions on Graphics (TOG)}, 42\penalty0 (6):\penalty0 1--12, 2023.

\bibitem[Kim et~al.(2024)Kim, Kim, Na, and Joo]{kim2024parahome}
Jeonghwan Kim, Jisoo Kim, Jeonghyeon Na, and Hanbyul Joo.
\newblock {ParaHome}: Parameterizing everyday home activities towards 3d generative modeling of human-object interactions.
\newblock \emph{arXiv preprint arXiv:2401.10232}, 2024.

\bibitem[Kim et~al.(2023)Kim, Saito, and Joo]{kim2023ncho}
Taeksoo Kim, Shunsuke Saito, and Hanbyul Joo.
\newblock {NCHO}: Unsupervised learning for neural 3d composition of humans and objects.
\newblock In \emph{ICCV}, 2023.

\bibitem[Krebs et~al.(2021)Krebs, Meixner, Patzer, and Asfour]{krebs2021kit}
Franziska Krebs, Andre Meixner, Isabel Patzer, and Tamim Asfour.
\newblock The kit bimanual manipulation dataset.
\newblock In \emph{Humanoids}, 2021.

\bibitem[Kulkarni et~al.(2023)Kulkarni, Rempe, Genova, Kundu, Johnson, Fouhey, and Guibas]{kulkarni2023nifty}
Nilesh Kulkarni, Davis Rempe, Kyle Genova, Abhijit Kundu, Justin Johnson, David Fouhey, and Leonidas Guibas.
\newblock {NIFTY}: Neural object interaction fields for guided human motion synthesis.
\newblock \emph{arXiv preprint arXiv:2307.07511}, 2023.

\bibitem[Lee and Joo(2023)]{lee2023locomotion}
Jiye Lee and Hanbyul Joo.
\newblock {Locomotion-Action-Manipulation}: Synthesizing human-scene interactions in complex 3d environments.
\newblock In \emph{ICCV}, 2023.

\bibitem[Li et~al.(2023{\natexlab{a}})Li, Clegg, Mottaghi, Wu, Puig, and Liu]{li2023controllable}
Jiaman Li, Alexander Clegg, Roozbeh Mottaghi, Jiajun Wu, Xavier Puig, and C~Karen Liu.
\newblock Controllable human-object interaction synthesis.
\newblock \emph{arXiv preprint arXiv:2312.03913}, 2023{\natexlab{a}}.

\bibitem[Li et~al.(2023{\natexlab{b}})Li, Wu, and Liu]{li2023object}
Jiaman Li, Jiajun Wu, and C~Karen Liu.
\newblock Object motion guided human motion synthesis.
\newblock \emph{ACM Transactions on Graphics (TOG)}, 42\penalty0 (6):\penalty0 1--11, 2023{\natexlab{b}}.

\bibitem[Li et~al.(2023{\natexlab{c}})Li, Wang, Loy, and Dai]{li2023task}
Quanzhou Li, Jingbo Wang, Chen~Change Loy, and Bo Dai.
\newblock Task-oriented human-object interactions generation with implicit neural representations.
\newblock \emph{arXiv preprint arXiv:2303.13129}, 2023{\natexlab{c}}.

\bibitem[Liang et~al.(2023)Liang, Zhang, Li, Yu, and Xu]{liang2023intergen}
Han Liang, Wenqian Zhang, Wenxuan Li, Jingyi Yu, and Lan Xu.
\newblock {InterGen}: Diffusion-based multi-human motion generation under complex interactions.
\newblock \emph{arXiv preprint arXiv:2304.05684}, 2023.

\bibitem[Lin et~al.(2023)Lin, Zeng, Lu, Cai, Zhang, Wang, and Zhang]{lin2023motionx}
Jing Lin, Ailing Zeng, Shunlin Lu, Yuanhao Cai, Ruimao Zhang, Haoqian Wang, and Lei Zhang.
\newblock {Motion-X}: A large-scale 3d expressive whole-body human motion dataset.
\newblock In \emph{NeurIPS}, 2023.

\bibitem[Liu and Hodgins(2018)]{liu2018learning}
Libin Liu and Jessica Hodgins.
\newblock Learning basketball dribbling skills using trajectory optimization and deep reinforcement learning.
\newblock \emph{ACM Transactions on Graphics (TOG)}, 37\penalty0 (4):\penalty0 1--14, 2018.

\bibitem[Liu et~al.(2023)Liu, Zhou, Yang, Gupta, and Wang]{liu2023contactgen}
Shaowei Liu, Yang Zhou, Jimei Yang, Saurabh Gupta, and Shenlong Wang.
\newblock Contactgen: Generative contact modeling for grasp generation.
\newblock In \emph{ICCV}, 2023.

\bibitem[Liu et~al.(2022)Liu, Liu, Jiang, Lyu, Wan, Shen, Liang, Fu, Wang, and Yi]{liu2022hoi4d}
Yunze Liu, Yun Liu, Che Jiang, Kangbo Lyu, Weikang Wan, Hao Shen, Boqiang Liang, Zhoujie Fu, He Wang, and Li Yi.
\newblock Hoi4d: A 4d egocentric dataset for category-level human-object interaction.
\newblock In \emph{CVPR}, 2022.

\bibitem[Loper et~al.(2015)Loper, Mahmood, Romero, Pons-Moll, and Black]{loper2015smpl}
Matthew Loper, Naureen Mahmood, Javier Romero, Gerard Pons-Moll, and Michael~J Black.
\newblock {SMPL}: A skinned multi-person linear model.
\newblock \emph{ACM transactions on graphics}, 2015.

\bibitem[Lu et~al.(2023)Lu, Chen, Zeng, Lin, Zhang, Zhang, and Shum]{humantomato}
Shunlin Lu, Ling-Hao Chen, Ailing Zeng, Jing Lin, Ruimao Zhang, Lei Zhang, and Heung-Yeung Shum.
\newblock Humantomato: Text-aligned whole-body motion generation.
\newblock \emph{arxiv:2310.12978}, 2023.

\bibitem[Luo et~al.(2024)Luo, Wang, Liu, Zhang, Tessler, Wang, Yuan, Cao, Lin, Wang, et~al.]{luo2024smplolympics}
Zhengyi Luo, Jiashun Wang, Kangni Liu, Haotian Zhang, Chen Tessler, Jingbo Wang, Ye Yuan, Jinkun Cao, Zihui Lin, Fengyi Wang, et~al.
\newblock Smplolympics: Sports environments for physically simulated humanoids.
\newblock \emph{arXiv preprint arXiv:2407.00187}, 2024.

\bibitem[Lv et~al.(2025)Lv, Xu, Yan, Jin, Xu, Wu, Liu, Li, Bi, Zeng, et~al.]{lv2025himo}
Xintao Lv, Liang Xu, Yichao Yan, Xin Jin, Congsheng Xu, Shuwen Wu, Yifan Liu, Lincheng Li, Mengxiao Bi, Wenjun Zeng, et~al.
\newblock Himo: A new benchmark for full-body human interacting with multiple objects.
\newblock In \emph{ECCV}, 2025.

\bibitem[Ma et~al.(2024)Ma, Xu, Chen, and Wang]{ma2024diff}
Junyi Ma, Jingyi Xu, Xieyuanli Chen, and Hesheng Wang.
\newblock Diff-ip2d: Diffusion-based hand-object interaction prediction on egocentric videos.
\newblock \emph{arXiv preprint arXiv:2405.04370}, 2024.

\bibitem[Mahmood et~al.(2019)Mahmood, Ghorbani, Troje, Pons-Moll, and Black]{mahmood2019amass}
Naureen Mahmood, Nima Ghorbani, Nikolaus~F Troje, Gerard Pons-Moll, and Michael~J Black.
\newblock {AMASS}: Archive of motion capture as surface shapes.
\newblock In \emph{ICCV}, 2019.

\bibitem[Makoviychuk et~al.(2021)Makoviychuk, Wawrzyniak, Guo, Lu, Storey, Macklin, Hoeller, Rudin, Allshire, Handa, et~al.]{makoviychuk2021isaac}
Viktor Makoviychuk, Lukasz Wawrzyniak, Yunrong Guo, Michelle Lu, Kier Storey, Miles Macklin, David Hoeller, Nikita Rudin, Arthur Allshire, Ankur Handa, et~al.
\newblock Isaac gym: High performance gpu-based physics simulation for robot learning.
\newblock \emph{arXiv preprint arXiv:2108.10470}, 2021.

\bibitem[Mandery et~al.(2015)Mandery, Terlemez, Do, Vahrenkamp, and Asfour]{Mandery2015a}
Christian Mandery, \"Omer Terlemez, Martin Do, Nikolaus Vahrenkamp, and Tamim Asfour.
\newblock The kit whole-body human motion database.
\newblock In \emph{ICAR}, 2015.

\bibitem[Mandery et~al.(2016)Mandery, Terlemez, Do, Vahrenkamp, and Asfour]{Mandery2016b}
Christian Mandery, \"Omer Terlemez, Martin Do, Nikolaus Vahrenkamp, and Tamim Asfour.
\newblock Unifying representations and large-scale whole-body motion databases for studying human motion.
\newblock \emph{IEEE Transactions on Robotics}, 32\penalty0 (4):\penalty0 796--809, 2016.

\bibitem[Mehta et~al.(2018)Mehta, Sotnychenko, Mueller, Xu, Sridhar, Pons-Moll, and Theobalt]{singleshotmultiperson2018}
Dushyant Mehta, Oleksandr Sotnychenko, Franziska Mueller, Weipeng Xu, Srinath Sridhar, Gerard Pons-Moll, and Christian Theobalt.
\newblock Single-shot multi-person {3D} pose estimation from monocular {RGB}.
\newblock In \emph{3DV}, 2018.

\bibitem[Merel et~al.(2020)Merel, Tunyasuvunakool, Ahuja, Tassa, Hasenclever, Pham, Erez, Wayne, and Heess]{merel2020catch}
Josh Merel, Saran Tunyasuvunakool, Arun Ahuja, Yuval Tassa, Leonard Hasenclever, Vu Pham, Tom Erez, Greg Wayne, and Nicolas Heess.
\newblock Catch \& carry: reusable neural controllers for vision-guided whole-body tasks.
\newblock \emph{ACM Transactions on Graphics (TOG)}, 39\penalty0 (4):\penalty0 39--1, 2020.

\bibitem[Moon et~al.(2020)Moon, Yu, Wen, Shiratori, and Lee]{moon2020interhand2}
Gyeongsik Moon, Shoou-I Yu, He Wen, Takaaki Shiratori, and Kyoung~Mu Lee.
\newblock Interhand2. 6m: A dataset and baseline for 3d interacting hand pose estimation from a single rgb image.
\newblock In \emph{ECCV}, 2020.

\bibitem[Nguyen et~al.(2024)Nguyen, Vu, Tran, and Nguyen]{nguyen2024dataset}
Quang Nguyen, Truong Vu, Anh Tran, and Khoi Nguyen.
\newblock Dataset diffusion: Diffusion-based synthetic data generation for pixel-level semantic segmentation.
\newblock In \emph{NeurIPS}, 2024.

\bibitem[Ohkawa et~al.(2023)Ohkawa, He, Sener, Hodan, Tran, and Keskin]{ohkawa2023assemblyhands}
Takehiko Ohkawa, Kun He, Fadime Sener, Tomas Hodan, Luan Tran, and Cem Keskin.
\newblock Assemblyhands: Towards egocentric activity understanding via 3d hand pose estimation.
\newblock In \emph{CVPR}, 2023.

\bibitem[Oord et~al.(2018)Oord, Li, and Vinyals]{oord2018representation}
Aaron van~den Oord, Yazhe Li, and Oriol Vinyals.
\newblock Representation learning with contrastive predictive coding.
\newblock \emph{arXiv preprint arXiv:1807.03748}, 2018.

\bibitem[OpenAI(2023)]{chatgpt}
OpenAI.
\newblock {ChatGPT}.
\newblock {\url{https://chat.openai.com/}}, 2023.

\bibitem[Pan et~al.(2023)Pan, Wang, Huang, Zhang, Wang, Tang, and Wang]{pan2023synthesizing}
Liang Pan, Jingbo Wang, Buzhen Huang, Junyu Zhang, Haofan Wang, Xu Tang, and Yangang Wang.
\newblock Synthesizing physically plausible human motions in 3d scenes.
\newblock \emph{arXiv preprint arXiv:2308.09036}, 2023.

\bibitem[Pavlakos et~al.(2019)Pavlakos, Choutas, Ghorbani, Bolkart, Osman, Tzionas, and Black]{SMPL-X:2019}
Georgios Pavlakos, Vasileios Choutas, Nima Ghorbani, Timo Bolkart, Ahmed A.~A. Osman, Dimitrios Tzionas, and Michael~J. Black.
\newblock Expressive body capture: {3D} hands, face, and body from a single image.
\newblock In \emph{CVPR}, 2019.

\bibitem[Peng et~al.(2023)Peng, Xie, Wu, Jampani, Sun, and Jiang]{peng2023hoi}
Xiaogang Peng, Yiming Xie, Zizhao Wu, Varun Jampani, Deqing Sun, and Huaizu Jiang.
\newblock {HOI-Diff}: Text-driven synthesis of 3d human-object interactions using diffusion models.
\newblock \emph{arXiv preprint arXiv:2312.06553}, 2023.

\bibitem[Petrov et~al.(2023)Petrov, Marin, Chibane, and Pons-Moll]{petrov2023object}
Ilya~A Petrov, Riccardo Marin, Julian Chibane, and Gerard Pons-Moll.
\newblock Object pop-up: Can we infer 3d objects and their poses from human interactions alone?
\newblock In \emph{CVPR}, 2023.

\bibitem[Petrovich et~al.(2022)Petrovich, Black, and Varol]{petrovich22temos}
Mathis Petrovich, Michael~J. Black, and G{\"u}l Varol.
\newblock {TEMOS}: Generating diverse human motions from textual descriptions.
\newblock In \emph{ECCV}, 2022.

\bibitem[Petrovich et~al.(2023)Petrovich, Black, and Varol]{petrovich2023tmr}
Mathis Petrovich, Michael~J Black, and G{\"u}l Varol.
\newblock {TMR}: Text-to-motion retrieval using contrastive 3d human motion synthesis.
\newblock In \emph{ICCV}, 2023.

\bibitem[Plappert et~al.(2016)Plappert, Mandery, and Asfour]{plappert2016kit}
Matthias Plappert, Christian Mandery, and Tamim Asfour.
\newblock The kit motion-language dataset.
\newblock \emph{Big data}, 4\penalty0 (4):\penalty0 236--252, 2016.

\bibitem[Prokudin et~al.(2019)Prokudin, Lassner, and Romero]{prokudin2019efficient}
Sergey Prokudin, Christoph Lassner, and Javier Romero.
\newblock Efficient learning on point clouds with basis point sets.
\newblock In \emph{ICCV}, 2019.

\bibitem[Punnakkal et~al.(2021)Punnakkal, Chandrasekaran, Athanasiou, Quiros-Ramirez, and Black]{BABEL:CVPR:2021}
Abhinanda~R. Punnakkal, Arjun Chandrasekaran, Nikos Athanasiou, Alejandra Quiros-Ramirez, and Michael~J. Black.
\newblock {BABEL}: Bodies, action and behavior with english labels.
\newblock In \emph{CVPR}, 2021.

\bibitem[Qi et~al.(2017)Qi, Yi, Su, and Guibas]{qi2017pointnet++}
Charles~Ruizhongtai Qi, Li Yi, Hao Su, and Leonidas~J Guibas.
\newblock Pointnet++: Deep hierarchical feature learning on point sets in a metric space.
\newblock In \emph{NeurIPS}, 2017.

\bibitem[Razali and Demiris(2023)]{razali2023action}
Haziq Razali and Yiannis Demiris.
\newblock Action-conditioned generation of bimanual object manipulation sequences.
\newblock In \emph{AAAI}, 2023.

\bibitem[Reimers and Gurevych(2019)]{reimers2019sentence}
Nils Reimers and Iryna Gurevych.
\newblock Sentence-bert: Sentence embeddings using siamese bert-networks.
\newblock \emph{arXiv preprint arXiv:1908.10084}, 2019.

\bibitem[Romero et~al.(2017)Romero, Tzionas, and Black]{MANO}
Javier Romero, Dimitrios Tzionas, and Michael~J. Black.
\newblock Embodied hands: Modeling and capturing hands and bodies together.
\newblock \emph{ACM Transactions on Graphics}, 36\penalty0 (6), 2017.

\bibitem[Song et~al.(2024)Song, Zhang, Li, Gao, Hao, Hou, Chen, Li, and Qin]{song2024hoianimator}
Wenfeng Song, Xinyu Zhang, Shuai Li, Yang Gao, Aimin Hao, Xia Hou, Chenglizhao Chen, Ning Li, and Hong Qin.
\newblock Hoianimator: Generating text-prompt human-object animations using novel perceptive diffusion models.
\newblock In \emph{CVPR}, 2024.

\bibitem[Starke et~al.(2019)Starke, Zhang, Komura, and Saito]{starke2019neural}
Sebastian Starke, He Zhang, Taku Komura, and Jun Saito.
\newblock Neural state machine for character-scene interactions.
\newblock \emph{ACM Trans. Graph.}, 38\penalty0 (6):\penalty0 209--1, 2019.

\bibitem[Starke et~al.(2020)Starke, Zhao, Komura, and Zaman]{starke2020local}
Sebastian Starke, Yiwei Zhao, Taku Komura, and Kazi Zaman.
\newblock Local motion phases for learning multi-contact character movements.
\newblock \emph{ACM Transactions on Graphics (TOG)}, 39\penalty0 (4):\penalty0 54--1, 2020.

\bibitem[Taheri et~al.(2020)Taheri, Ghorbani, Black, and Tzionas]{taheri2020grab}
Omid Taheri, Nima Ghorbani, Michael~J Black, and Dimitrios Tzionas.
\newblock {GRAB}: A dataset of whole-body human grasping of objects.
\newblock In \emph{ECCV}, 2020.

\bibitem[Taheri et~al.(2022)Taheri, Choutas, Black, and Tzionas]{taheri2022goal}
Omid Taheri, Vasileios Choutas, Michael~J Black, and Dimitrios Tzionas.
\newblock {GOAL}: Generating 4d whole-body motion for hand-object grasping.
\newblock In \emph{CVPR}, 2022.

\bibitem[Taheri et~al.(2024)Taheri, Zhou, Tzionas, Zhou, Ceylan, Pirk, and Black]{taheri2024grip}
Omid Taheri, Yi Zhou, Dimitrios Tzionas, Yang Zhou, Duygu Ceylan, Soren Pirk, and Michael~J Black.
\newblock Grip: Generating interaction poses using spatial cues and latent consistency.
\newblock In \emph{3DV}, 2024.

\bibitem[Tendulkar et~al.(2023)Tendulkar, Sur{\'\i}s, and Vondrick]{tendulkar2023flex}
Purva Tendulkar, D{\'\i}dac Sur{\'\i}s, and Carl Vondrick.
\newblock Flex: Full-body grasping without full-body grasps.
\newblock In \emph{ICCV}, 2023.

\bibitem[Tessler et~al.(2024)Tessler, Guo, Nabati, Chechik, and Peng]{tessler2024maskedmimic}
Chen Tessler, Yunrong Guo, Ofir Nabati, Gal Chechik, and Xue~Bin Peng.
\newblock Maskedmimic: Unified physics-based character control through masked motion inpainting.
\newblock \emph{arXiv preprint arXiv:2409.14393}, 2024.

\bibitem[Tevet et~al.(2022)Tevet, Raab, Gordon, Shafir, Cohen-Or, and Bermano]{tevet2022human}
Guy Tevet, Sigal Raab, Brian Gordon, Yonatan Shafir, Daniel Cohen-Or, and Amit~H Bermano.
\newblock Human motion diffusion model.
\newblock \emph{arXiv preprint arXiv:2209.14916}, 2022.

\bibitem[Tevet et~al.(2024)Tevet, Raab, Cohan, Reda, Luo, Peng, Bermano, and van~de Panne]{tevet2024closd}
Guy Tevet, Sigal Raab, Setareh Cohan, Daniele Reda, Zhengyi Luo, Xue~Bin Peng, Amit~H Bermano, and Michiel van~de Panne.
\newblock Closd: Closing the loop between simulation and diffusion for multi-task character control.
\newblock \emph{arXiv preprint arXiv:2410.03441}, 2024.

\bibitem[Tian et~al.(2024)Tian, Yang, Ji, Ma, Xu, Yu, Shi, and Wang]{tian2024gaze}
Jie Tian, Lingxiao Yang, Ran Ji, Yuexin Ma, Lan Xu, Jingyi Yu, Ye Shi, and Jingya Wang.
\newblock Gaze-guided hand-object interaction synthesis: Benchmark and method.
\newblock \emph{arXiv preprint arXiv:2403.16169}, 2024.

\bibitem[Wan et~al.(2022)Wan, Yang, Liu, Zhang, Jia, Choi, Pan, Theobalt, Komura, and Wang]{9714029}
Weilin Wan, Lei Yang, Lingjie Liu, Zhuoying Zhang, Ruixing Jia, Yi-King Choi, Jia Pan, Christian Theobalt, Taku Komura, and Wenping Wang.
\newblock Learn to predict how humans manipulate large-sized objects from interactive motions.
\newblock \emph{IEEE Robotics and Automation Letters}, 2022.

\bibitem[Wang et~al.(2024{\natexlab{a}})Wang, Hodgins, and Won]{wang2024strategy}
Jiashun Wang, Jessica Hodgins, and Jungdam Won.
\newblock Strategy and skill learning for physics-based table tennis animation.
\newblock In \emph{SIGGRAPH}, 2024{\natexlab{a}}.

\bibitem[Wang et~al.(2022)Wang, Li, Kuo, Kocabas, Aksan, and Hilliges]{wang2022reconstructing}
Xi Wang, Gen Li, Yen-Ling Kuo, Muhammed Kocabas, Emre Aksan, and Otmar Hilliges.
\newblock Reconstructing action-conditioned human-object interactions using commonsense knowledge priors.
\newblock In \emph{3DV}, 2022.

\bibitem[Wang et~al.(2023{\natexlab{a}})Wang, Lin, Zeng, Luo, Zhang, and Zhang]{wang2023physhoi}
Yinhuai Wang, Jing Lin, Ailing Zeng, Zhengyi Luo, Jian Zhang, and Lei Zhang.
\newblock {PhysHOI}: Physics-based imitation of dynamic human-object interaction.
\newblock \emph{arXiv preprint arXiv:2312.04393}, 2023{\natexlab{a}}.

\bibitem[Wang et~al.(2024{\natexlab{b}})Wang, Zhao, Yu, Zeng, Lin, Luo, Tsui, Yu, Li, Chen, et~al.]{wang2024skillmimic}
Yinhuai Wang, Qihan Zhao, Runyi Yu, Ailing Zeng, Jing Lin, Zhengyi Luo, Hok~Wai Tsui, Jiwen Yu, Xiu Li, Qifeng Chen, et~al.
\newblock Skillmimic: Learning reusable basketball skills from demonstrations.
\newblock \emph{arXiv preprint arXiv:2408.15270}, 2024{\natexlab{b}}.

\bibitem[Wang et~al.(2023{\natexlab{b}})Wang, Wang, Lin, and Dai]{wang2023intercontrol}
Zhenzhi Wang, Jingbo Wang, Dahua Lin, and Bo Dai.
\newblock {InterControl}: Generate human motion interactions by controlling every joint.
\newblock \emph{arXiv preprint arXiv:2311.15864}, 2023{\natexlab{b}}.

\bibitem[Wiederhold et~al.(2024)Wiederhold, Megyeri, Paris, Banerjee, and Banerjee]{wiederhold2024hoh}
Noah Wiederhold, Ava Megyeri, DiMaggio Paris, Sean Banerjee, and Natasha Banerjee.
\newblock Hoh: Markerless multimodal human-object-human handover dataset with large object count.
\newblock In \emph{NeurIPS}, 2024.

\bibitem[Wu et~al.(2023)Wu, Escontrela, Hafner, Abbeel, and Goldberg]{wu2023daydreamer}
Philipp Wu, Alejandro Escontrela, Danijar Hafner, Pieter Abbeel, and Ken Goldberg.
\newblock Daydreamer: World models for physical robot learning.
\newblock In \emph{CoRL}, 2023.

\bibitem[Wu et~al.(2024{\natexlab{a}})Wu, Shi, Huang, Yu, Xu, and Wang]{wu2024thor}
Qianyang Wu, Ye Shi, Xiaoshui Huang, Jingyi Yu, Lan Xu, and Jingya Wang.
\newblock {THOR}: Text to human-object interaction diffusion via relation intervention.
\newblock \emph{arXiv preprint arXiv:2403.11208}, 2024{\natexlab{a}}.

\bibitem[Wu et~al.(2022)Wu, Wang, Zhang, Zhang, Hilliges, Yu, and Tang]{wu2022saga}
Yan Wu, Jiahao Wang, Yan Zhang, Siwei Zhang, Otmar Hilliges, Fisher Yu, and Siyu Tang.
\newblock {SAGA}: Stochastic whole-body grasping with contact.
\newblock In \emph{ECCV}, 2022.

\bibitem[Wu et~al.(2024{\natexlab{b}})Wu, Li, and Liu]{wu2024human}
Zhen Wu, Jiaman Li, and C~Karen Liu.
\newblock Human-object interaction from human-level instructions.
\newblock \emph{arXiv preprint arXiv:2406.17840}, 2024{\natexlab{b}}.

\bibitem[Xie et~al.(2022{\natexlab{a}})Xie, Bhatnagar, and Pons-Moll]{xie2022chore}
Xianghui Xie, Bharat~Lal Bhatnagar, and Gerard Pons-Moll.
\newblock Chore: Contact, human and object reconstruction from a single rgb image.
\newblock In \emph{ECCV}, 2022{\natexlab{a}}.

\bibitem[Xie et~al.(2024)Xie, Lenssen, and Pons-Moll]{xie2024intertrack}
Xianghui Xie, Jan~Eric Lenssen, and Gerard Pons-Moll.
\newblock {InterTrack}: Tracking human object interaction without object templates.
\newblock \emph{arXiv preprint arXiv:2408.13953}, 2024.

\bibitem[Xie et~al.(2023{\natexlab{a}})Xie, Jampani, Zhong, Sun, and Jiang]{xie2023omnicontrol}
Yiming Xie, Varun Jampani, Lei Zhong, Deqing Sun, and Huaizu Jiang.
\newblock {OmniControl}: Control any joint at any time for human motion generation.
\newblock \emph{arXiv preprint arXiv:2310.08580}, 2023{\natexlab{a}}.

\bibitem[Xie et~al.(2022{\natexlab{b}})Xie, Starke, Ling, and van~de Panne]{xie2022learning}
Zhaoming Xie, Sebastian Starke, Hung~Yu Ling, and Michiel van~de Panne.
\newblock Learning soccer juggling skills with layer-wise mixture-of-experts.
\newblock In \emph{SIGGRAPH}, 2022{\natexlab{b}}.

\bibitem[Xie et~al.(2023{\natexlab{b}})Xie, Tseng, Starke, van~de Panne, and Liu]{xie2023hierarchical}
Zhaoming Xie, Jonathan Tseng, Sebastian Starke, Michiel van~de Panne, and C~Karen Liu.
\newblock Hierarchical planning and control for box loco-manipulation.
\newblock \emph{arXiv preprint arXiv:2306.09532}, 2023{\natexlab{b}}.

\bibitem[Xu et~al.(2023{\natexlab{a}})Xu, Lv, Yan, Jin, Wu, Xu, Liu, Zhou, Rao, Sheng, et~al.]{xu2023inter}
Liang Xu, Xintao Lv, Yichao Yan, Xin Jin, Shuwen Wu, Congsheng Xu, Yifan Liu, Yizhou Zhou, Fengyun Rao, Xingdong Sheng, et~al.
\newblock Inter-x: Towards versatile human-human interaction analysis.
\newblock \emph{arXiv preprint arXiv:2312.16051}, 2023{\natexlab{a}}.

\bibitem[Xu et~al.(2023{\natexlab{b}})Xu, Li, Wang, and Gui]{xu2023interdiff}
Sirui Xu, Zhengyuan Li, Yu-Xiong Wang, and Liang-Yan Gui.
\newblock {InterDiff}: Generating 3d human-object interactions with physics-informed diffusion.
\newblock In \emph{ICCV}, 2023{\natexlab{b}}.

\bibitem[Xu et~al.(2024)Xu, Wang, Wang, and Gui]{xu2024interdreamer}
Sirui Xu, Ziyin Wang, Yu-Xiong Wang, and Liang-Yan Gui.
\newblock Interdreamer: Zero-shot text to 3d dynamic human-object interaction.
\newblock \emph{arXiv preprint arXiv:2403.19652}, 2024.

\bibitem[Xu et~al.(2025)Xu, Ling, Wang, and Gui]{xu2025intermimic}
Sirui Xu, Hung~Yu Ling, Yu-Xiong Wang, and Liang-Yan Gui.
\newblock {InterMimic}: Towards universal whole-body control for physics-based human-object interactions.
\newblock In \emph{CVPR}, 2025.

\bibitem[Xu et~al.(2021)Xu, Joo, Mori, and Savva]{xu2021d3dhoi}
Xiang Xu, Hanbyul Joo, Greg Mori, and Manolis Savva.
\newblock {D3D-HOI}: Dynamic 3d human-object interactions from videos.
\newblock \emph{arXiv preprint arXiv:2108.08420}, 2021.

\bibitem[Yang et~al.(2024{\natexlab{a}})Yang, Kang, Kong, Oh, and Kang]{yang2024person}
ChangHee Yang, ChanHee Kang, Kyeongbo Kong, Hanni Oh, and Suk-Ju Kang.
\newblock Person in place: Generating associative skeleton-guidance maps for human-object interaction image editing.
\newblock In \emph{CVPR}, 2024{\natexlab{a}}.

\bibitem[Yang et~al.(2024{\natexlab{b}})Yang, Niu, Jiang, Zhang, and Huang]{yang2024f}
Jie Yang, Xuesong Niu, Nan Jiang, Ruimao Zhang, and Siyuan Huang.
\newblock {F-HOI}: Toward fine-grained semantic-aligned 3d human-object interactions.
\newblock In \emph{ECCV}, 2024{\natexlab{b}}.

\bibitem[Yang et~al.(2024{\natexlab{c}})Yang, Zhai, Luo, Cao, and Zha]{yang2024lemon}
Yuhang Yang, Wei Zhai, Hongchen Luo, Yang Cao, and Zheng-Jun Zha.
\newblock Lemon: Learning 3d human-object interaction relation from 2d images.
\newblock In \emph{CVPR}, 2024{\natexlab{c}}.

\bibitem[Yang et~al.(2022)Yang, Yin, and Liu]{yang2022learning}
Zeshi Yang, Kangkang Yin, and Libin Liu.
\newblock Learning to use chopsticks in diverse gripping styles.
\newblock \emph{ACM Transactions on Graphics (TOG)}, 41\penalty0 (4):\penalty0 1--17, 2022.

\bibitem[Yao et~al.(2024)Yao, Song, Zhou, Ao, Chen, and Liu]{yao2024moconvq}
Heyuan Yao, Zhenhua Song, Yuyang Zhou, Tenglong Ao, Baoquan Chen, and Libin Liu.
\newblock Moconvq: Unified physics-based motion control via scalable discrete representations.
\newblock \emph{ACM Transactions on Graphics (TOG)}, 43\penalty0 (4):\penalty0 1--21, 2024.

\bibitem[Ye et~al.(2023)Ye, Li, Gupta, De~Mello, Birchfield, Song, Tulsiani, and Liu]{ye2023affordance}
Yufei Ye, Xueting Li, Abhinav Gupta, Shalini De~Mello, Stan Birchfield, Jiaming Song, Shubham Tulsiani, and Sifei Liu.
\newblock Affordance diffusion: Synthesizing hand-object interactions.
\newblock In \emph{CVPR}, 2023.

\bibitem[Zhan et~al.(2024)Zhan, Yang, Zhao, Mao, Xu, Lin, Li, and Lu]{zhan2024oakink2}
Xinyu Zhan, Lixin Yang, Yifei Zhao, Kangrui Mao, Hanlin Xu, Zenan Lin, Kailin Li, and Cewu Lu.
\newblock Oakink2: A dataset of bimanual hands-object manipulation in complex task completion.
\newblock \emph{arXiv preprint arXiv:2403.19417}, 2024.

\bibitem[Zhang et~al.(2024{\natexlab{a}})Zhang, Liu, Xing, Tang, and Yi]{zhang2024core4d}
Chengwen Zhang, Yun Liu, Ruofan Xing, Bingda Tang, and Li Yi.
\newblock Core4d: A 4d human-object-human interaction dataset for collaborative object rearrangement.
\newblock \emph{arXiv preprint arXiv:2406.19353}, 2024{\natexlab{a}}.

\bibitem[Zhang et~al.(2023{\natexlab{a}})Zhang, Christen, Fan, Zheng, Hwangbo, Song, and Hilliges]{zhang2023artigrasp}
Hui Zhang, Sammy Christen, Zicong Fan, Luocheng Zheng, Jemin Hwangbo, Jie Song, and Otmar Hilliges.
\newblock {ArtiGrasp}: Physically plausible synthesis of bi-manual dexterous grasping and articulation.
\newblock \emph{arXiv preprint arXiv:2309.03891}, 2023{\natexlab{a}}.

\bibitem[Zhang et~al.(2023{\natexlab{b}})Zhang, Luo, Yang, Xu, Wu, Shi, Yu, Xu, and Wang]{zhang2023neuraldome}
Juze Zhang, Haimin Luo, Hongdi Yang, Xinru Xu, Qianyang Wu, Ye Shi, Jingyi Yu, Lan Xu, and Jingya Wang.
\newblock {NeuralDome}: A neural modeling pipeline on multi-view human-object interactions.
\newblock In \emph{CVPR}, 2023{\natexlab{b}}.

\bibitem[Zhang et~al.(2023{\natexlab{c}})Zhang, Zhang, Cun, Huang, Zhang, Zhao, Lu, and Shen]{zhang2023t2m}
Jianrong Zhang, Yangsong Zhang, Xiaodong Cun, Shaoli Huang, Yong Zhang, Hongwei Zhao, Hongtao Lu, and Xi Shen.
\newblock {T2M-GPT}: Generating human motion from textual descriptions with discrete representations.
\newblock In \emph{CVPR}, 2023{\natexlab{c}}.

\bibitem[Zhang et~al.(2024{\natexlab{b}})Zhang, Zhang, Song, Shi, Zhao, Shi, Yu, Xu, and Wang]{zhang2024hoi}
Juze Zhang, Jingyan Zhang, Zining Song, Zhanhe Shi, Chengfeng Zhao, Ye Shi, Jingyi Yu, Lan Xu, and Jingya Wang.
\newblock Hoi-m\^{} 3: Capture multiple humans and objects interaction within contextual environment.
\newblock In \emph{CVPR}, 2024{\natexlab{b}}.

\bibitem[Zhang et~al.(2024{\natexlab{c}})Zhang, Zhang, An, Li, Zhang, Hu, and Liu]{zhang2024manidext}
Jiajun Zhang, Yuxiang Zhang, Liang An, Mengcheng Li, Hongwen Zhang, Zonghai Hu, and Yebin Liu.
\newblock Manidext: Hand-object manipulation synthesis via continuous correspondence embeddings and residual-guided diffusion.
\newblock \emph{arXiv preprint arXiv:2409.09300}, 2024{\natexlab{c}}.

\bibitem[Zhang et~al.(2020)Zhang, Pepose, Joo, Ramanan, Malik, and Kanazawa]{zhang2020perceiving}
Jason~Y Zhang, Sam Pepose, Hanbyul Joo, Deva Ramanan, Jitendra Malik, and Angjoo Kanazawa.
\newblock Perceiving 3d human-object spatial arrangements from a single image in the wild.
\newblock In \emph{ECCV}, 2020.

\bibitem[Zhang et~al.(2022)Zhang, Bhatnagar, Starke, Guzov, and Pons-Moll]{zhang2022couch}
Xiaohan Zhang, Bharat~Lal Bhatnagar, Sebastian Starke, Vladimir Guzov, and Gerard Pons-Moll.
\newblock {COUCH}: Towards controllable human-chair interactions.
\newblock In \emph{ECCV}, 2022.

\bibitem[Zhang et~al.(2024{\natexlab{d}})Zhang, Bhatnagar, Starke, Petrov, Guzov, Dhamo, P{\'e}rez-Pellitero, and Pons-Moll]{zhang2024force}
Xiaohan Zhang, Bharat~Lal Bhatnagar, Sebastian Starke, Ilya Petrov, Vladimir Guzov, Helisa Dhamo, Eduardo P{\'e}rez-Pellitero, and Gerard Pons-Moll.
\newblock {FORCE}: Dataset and method for intuitive physics guided human-object interaction.
\newblock \emph{arXiv preprint arXiv:2403.11237}, 2024{\natexlab{d}}.

\bibitem[Zhang et~al.(2023{\natexlab{d}})Zhang, Gopinath, Ye, Hodgins, Turk, and Won]{zhang2023simulation}
Yunbo Zhang, Deepak Gopinath, Yuting Ye, Jessica Hodgins, Greg Turk, and Jungdam Won.
\newblock Simulation and retargeting of complex multi-character interactions.
\newblock In \emph{SIGGRAPH}, 2023{\natexlab{d}}.

\bibitem[Zhao et~al.(2024)Zhao, Zhang, Du, Shan, Wang, Yu, Wang, and Xu]{zhao2023im}
Chengfeng Zhao, Juze Zhang, Jiashen Du, Ziwei Shan, Junye Wang, Jingyi Yu, Jingya Wang, and Lan Xu.
\newblock {I'M HOI}: Inertia-aware monocular capture of 3d human-object interactions.
\newblock In \emph{CVPR}, 2024.

\bibitem[Zhao et~al.(2022)Zhao, Wang, Zhang, Beeler, and Tang]{zhao2022compositional}
Kaifeng Zhao, Shaofei Wang, Yan Zhang, Thabo Beeler, and Siyu Tang.
\newblock Compositional human-scene interaction synthesis with semantic control.
\newblock In \emph{ECCV}, 2022.

\bibitem[Zhao et~al.(2023)Zhao, Zhang, Wang, Beeler, and Tang]{Zhao:ICCV:2023}
Kaifeng Zhao, Yan Zhang, Shaofei Wang, Thabo Beeler, and Siyu Tang.
\newblock Synthesizing diverse human motions in 3d indoor scenes.
\newblock In \emph{ICCV}, 2023.

\bibitem[Zheng et~al.(2023)Zheng, Zheng, Fang, Liu, and Yi]{zheng2023cams}
Juntian Zheng, Qingyuan Zheng, Lixing Fang, Yun Liu, and Li Yi.
\newblock {CAMS}: Canonicalized manipulation spaces for category-level functional hand-object manipulation synthesis.
\newblock In \emph{CVPR}, 2023.

\bibitem[Zhou et~al.(2022)Zhou, Bhatnagar, Lenssen, and Pons-Moll]{zhou2022toch}
Keyang Zhou, Bharat~Lal Bhatnagar, Jan~Eric Lenssen, and Gerard Pons-Moll.
\newblock Toch: Spatio-temporal object-to-hand correspondence for motion refinement.
\newblock In \emph{ECCV}, 2022.

\end{thebibliography}
}
\clearpage
\setcounter{page}{1}
\maketitlesupplementary

\setcounter{table}{0}
\renewcommand{\thetable}{\Alph{table}}
\renewcommand*{\theHtable}{\thetable}
\setcounter{figure}{0}
\renewcommand{\thefigure}{\Alph{figure}}
\renewcommand*{\theHfigure}{\thefigure}
\setcounter{section}{0}
\renewcommand{\thesection}{\Alph{section}}
\renewcommand*{\theHsection}{\thesection}

\noindent We will release the complete dataset including the consolidated MoCap data and our synthetic data, as well as models for benchmarking tasks. In this supplementary material, we introduce: (\textbf{1}) a \textbf{website featuring demo videos} of benchmark tasks, demonstrating how interaction correction and augmentation improve data quality and scale; (\textbf{2}) additional illustrations and details of our data curation, correction, and augmentation processes in Sec.~\ref{sec:method_supp}; (\textbf{3}) further implementation details and experimental results in Sec.~\ref{sec:exp_supp}, which were omitted from the main paper due to space limitations; (\textbf{4}) licensing information in Sec.~\ref{sec:license}; and (\textbf{5}) a discussion of limitations and potential negative societal impacts of our work in Sec.~\ref{sec:ethics}.

\section{Webpage} \label{sec:dataset}

\subsection{Webpage}
Our website is at \url{https://sirui-xu.github.io/InterAct}. It includes demo videos comparing the raw data with our corrected and augmented versions, as well as showcasing generated results across six interaction generation tasks. These demos complement the quantitative results in the main paper, demonstrating that our unified multi-task model, trained on our extensive dataset, achieves state-of-the-art performance. We also compare our marker-based representation to joint position and rotation-based representations, demonstrating that markers as representative surface vertices are better suited for interaction modeling, since contact occurs on surfaces rather than joints.

\section{Data Collection, Annotation, and Unification}\label{sec:method_supp}

\subsection{Text and Action Annotation}

In Sec.~\ref{sec:collection} of the main paper, we describe our process for manually annotating interaction sequences with detailed text instructions. Here, we elaborate on how we leverage large language models for automatic annotation augmentation, and generate additional action labels that facilitate further tasks.

We use GPT-4~\cite{chatgpt} to rephrase, simplify, and generate action labels for the annotations we collected. Specifically, for rephrasing and simplification, we send two messages to the API: a system message containing the requirements and the annotation that needs to be processed. For example, the system message we use for the simplification task is:
\begin{quote}
\centering
“I will provide a few sentences describing a human interacting with an object. Your task is to shorten the description while retaining its meaning, ensuring the object's name remains unchanged.
\end{quote}

To generate action labels, we prompt GPT-4 to identify from 15 predefined action labels and categorize the annotations accordingly. Specifically, the prompt we use is:
\begin{quote}
\centering
"I will provide a few sentences describing a human interacting with an object, and you need to select the single most fitting word to describe interactions from the following set: [Carry, Sit, Swing, Exercise, Rotate, Move, Hold, Drink, Eat, Play, Adjust, Lift, Kick, Pass, Manipulate]. Your response should be one word only."
\end{quote}
We include some of our manually annotated action labels in the prompt to enhance in-context learning and response accuracy of the language model.

\subsection{Processing of Each Sub-dataset}
\noindent \textbf{GRAB}~\cite{taheri2020grab} provides full 3D body shape, pose, and 3D object pose data captured using MoCap markers. We utilize their SMPL-X~\cite{SMPL-X:2019} human annotation, downsample interaction sequences from 120 to 30 fps, and align the data to our unified global coordinate system and ground heights.

\noindent \textbf{BEHAVE}~\cite{bhatnagar22behave} contains HOI video frames captured from multi-view RGBD sequences. We use their SMPL-H~\cite{MANO} human annotation. We align their data into our unified global coordinate system and ground heights.

\noindent \textbf{InterCap}~\cite{huang2022intercap}
contains HOI video frames captured from multi-view RGBD sequences. We use their SMPL-X~\cite{SMPL-X:2019} human annotation. We align their data into our unified global coordinate system and ground heights.

\noindent \textbf{Chairs}~\cite{jiang2022chairs} 
contains HOI video frames captured from multi-view RGBD sequences. We only select sequences that contain rigid objects. We only select sequences that contain rigid objects. To correct the tilted human and object, we calculate the ground normal vector using the lowest point set of the human and object in each frame. We use their SMPL-X~\cite{SMPL-X:2019} human annotation, interpolating interaction sequences from 10 to 30 fps. We align their data into our unified global coordinate system and ground heights.

\noindent \textbf{HODome}~\cite{zhang2023neuraldome} contains a total of HOI video frames captured from 76 viewpoints. The human body data are captured from these multi-view images using EasyMocap~\cite{easymocap}. We further process these representations to standard SMPL-H~\cite{MANO} annotation, downsample interaction sequences from 60 to 30 fps, and align them into the same global coordinate system and ground heights.

\noindent \textbf{OMOMO}~\cite{li2023object}
contains object and human motion captured using a Vicon system comprised of 12 cameras controlled by Vicon Shogun. We use their SMPL-X~\cite{SMPL-X:2019} human annotation. We align their data into our unified global coordinate system and ground heights.

\noindent \textbf{IMHD}~\cite{zhao2023im}
contains video frames captured from both RGB cameras and the object-mounted Inertial Measurement Unit (IMU). We process their human representations to standard SMPL-H~\cite{MANO} annotation, downsample interaction sequences from 60 to 30 fps, and align them into the same global coordinate system and ground heights.

\subsection{Interaction Correction} \label{sec:correction_supp}
In this section, we provide the formulation or explanation of learning objectives for interaction correction and augmentation, which are omitted from Sec.~\ref{sec:correction} of the main paper due to space constraints. 

\noindent \textbf{Hand Correction.}  
We define the hand poses as \(\{\boldsymbol{\mathrm{hand}}_i\}_{i=1}^L\) of arbitrary length \(L\). The learning objectives are,

\noindent(\textbf{1}) \textit{Penetration Loss}. Given the signed distance field of the human $\textbf{sdf}_i$, we employ a penetration loss to penalize the body-object interpenetration,
\begin{align}
    E_{\mathrm{pene}} = -\sum_{i=1}^{L}\sum_{d_o}\min(\textbf{sdf}_i(\boldsymbol{v}_{\boldsymbol o_i}[k]), 0),
\end{align}
where $\boldsymbol{v}_{\boldsymbol o_i}[k]$ refers to the object vertex of index $k$ at frame i.

\noindent(\textbf{2}) \textit{Smooth Loss.} We employ a smoothing loss to avoid excessive speed and acceleration changes.
\begin{align}  
    E_{\mathrm{smooth}} = &\sum_{i=1}^{L-1}\|{\boldsymbol {\mathrm{hand}}_{i+1}}- {\boldsymbol {\mathrm{hand}}_{i}}\|_2^2\notag\\+
    \sum_{i=1}^{L-2}\|(&{\boldsymbol {\mathrm{hand}}_{i+2}}- {\boldsymbol {\mathrm{hand}}_{i+1}})-({\boldsymbol {\mathrm{hand}}_{i+1}}- {\boldsymbol {\mathrm{hand}}_{i}})\|_2^2 
\end{align}
where $\boldsymbol {\mathrm{hand}}_{i}$ represents the hand pose at time step $i$

\noindent(\textbf{3}) \textit{Prior Loss.} We apply a prior loss to maintain natural hand poses and prevent the hand joints from exceeding their range of motion (RoM) due to excessive guidance from the Contact Loss. We set the RoM constraints as \(({\boldsymbol {\mathrm{hand}}_\mathrm{max}}, {\boldsymbol {\mathrm{hand}}_\mathrm{min}})\) for all joint, derived from the statistical analysis of the GRAB~\cite{taheri2020grab} dataset. The loss is defined as,
\begin{align}
    E_{\mathrm{prior}} &= \sum_{i=1}^{L}\|\min({\boldsymbol {\mathrm{hand}}_{i}}-{\boldsymbol {\mathrm{hand}}_\mathrm{min}},0)\|_2^2 \notag \\
     &+\|\max({\boldsymbol {\mathrm{hand}}_{i}}-{\boldsymbol {\mathrm{hand}}_\mathrm{max}},0)\|_2^2
\end{align}

\noindent\textit{Hyperparameters of Contact Promotion.} We include the formulation of the contact indicator ${c}_i$ for contact promotion loss defined in Sec.~\ref{sec:correction} of the main paper.
\begin{equation}
{c}_i = 
\begin{cases}
1 & \min_j d_j[i] \leq \epsilon \\
0 & \min_j d_j[i] > \epsilon_2 \\
-12.5 \min_j d_j[i] + 1.25 & \text{otherwise}, 
\end{cases}
\end{equation}
where $\min_j d_j[i]$ refers to hand-object chamfer distance, $\epsilon=0.02$ indicates the contact threshold following~\cite{bhatnagar22behave}, and $\epsilon_2=0.10$ indicates the non-contact threshold. The expression \( -12.5 \min_j d_j[i] + 1.25 \) provides the linear interpolation between these two phases. 

\noindent \textbf{Full-Body Correction.} In full-body correction,
we jointly optimize the full-body human pose \(\{\boldsymbol{h}_i\}_{i=1}^L\) and the object pose \(\{\boldsymbol{o}_i\}_{i=1}^L\), given the ground truth counterparts \(\{\boldsymbol{h}_i^\ast\}_{i=1}^L\) and \(\{\boldsymbol{o}_i^\ast\}_{i=1}^L\).
The overall objective is defined as,
\begin{align}
    E =   \lambda_{\mathrm{pene}}E_{\mathrm{pene}}+\lambda_{\mathrm{smooth}}E_{\mathrm{smooth}}+\lambda_{\mathrm{rec}}E_{\mathrm{rec}},
\end{align}
where each component of the loss is defined as follows:

\noindent (\textbf{1}) \textit{Penetration Loss.} Same as defined for hand correction.

\noindent (\textbf{2}) \textit{Smooth Loss.} The smooth loss incorporates additional terms for object motion.

\begin{align}
    E_{\mathrm{s}} = \sum_{i=1}^{L-1}\|{\boldsymbol h_{i+1}}- {\boldsymbol h_{i}}\|_2^2+ +\sum_{i=1}^{L-1}\|{\boldsymbol o_{i+1}}- {\boldsymbol o_{i}}\|_2^2 \notag \\ + \sum_{i=1}^{L-2}\|({\boldsymbol h_{i+2}}- {\boldsymbol h_{i+1}})-({\boldsymbol h_{i+1}}- {\boldsymbol h_{i}})\|_2^2 \notag \\ 
     + \sum_{i=1}^{L-2}\|({\boldsymbol o_{i+2}}- {\boldsymbol o_{i+1}})-({\boldsymbol o_{i+1}}- {\boldsymbol o_{i}})\|_2^2 
\end{align}

\noindent (\textbf{3}) \textit{Reconstruction Loss.} The reconstruction loss promote the optimized human and object pose to be close to the ground truth ${\boldsymbol h_{i}^{\ast}}$ and ${\boldsymbol o_{i}^{\ast}}$,
\begin{align}
    E_{\mathrm{rec}} = &\sum_{i=1}^{L}\|{\boldsymbol h_{i}}- {\boldsymbol h_{i}^{\ast}}\| + \|{\boldsymbol o_{i}}- {\boldsymbol o_{i}^{\ast}}\|
\end{align}

\subsection{Interaction Augmentation}

We align the human body to maintain interaction with the object to achieve contact invariance as we highlight in Sec.~\ref{sec:correction} of the main paper. In addition to the learning objective (\textbf{1}) \( E_{\mathrm{align}} \), introduced in the main paper to ensure contact consistency, we incorporate two additional objectives: (\textbf{2}) \( E_{\mathrm{reg}} \), a regularization term that penalizes excessive deviations from the original human body pose, with a particular focus on key body parts not in contact with the object; and (\textbf{3}) \( E_{\mathrm{smooth}} \), as defined in Sec.~\ref{sec:correction_supp}, which promotes temporal smoothness between frames by restricting velocity and acceleration. The regularization term is derived from the reconstruction loss described in Sec.~\ref{sec:correction_supp}, and is reformulated as follows:

\[
E_{\mathrm{reg}} = \beta \sum_{i=1}^{L} m_{m}\|{\boldsymbol h_{i}} - {\boldsymbol h_{i}^{\ast}}\| + \frac{1}{\beta} \sum_{i=1}^{L} \|{\boldsymbol h_{i}} - {\boldsymbol h_{i}^{\ast}}\|,
\]
where \( m_{m} \) is a mask applied to joints that are not involved in the interaction, and \( \beta = 5 \) is selected to emphasize these vertices.
The overall objective for augmentation is then defined as:

\begin{align}
    E_{\mathrm{aug}} = E_{\mathrm{align}} + E_{\mathrm{reg}} + E_{\mathrm{s}},
\end{align}
where the optimization runs for 300 iterations to update the human body pose \( \boldsymbol h \).

\section{Additional Implementation Details and Experimental Analysis} \label{sec:exp_supp}

\subsection{Marker-Based Representation with Shape Variance}
Our method uses the marker-based representation that couples pose and shape. Despite the coupled representation, our model can generate diverse human shapes during interaction generation. Although our task does not focus on specific human shape control and our text descriptions do not specify detailed characteristics like height or body type, the model inherently captures variability from the training data. This results in diverse human shapes, with heights varying from 1.71m to 1.81m across 50 batches of generation. 

\subsection{Text-Conditioned Interaction Generation} 
\noindent\textbf{Additional Implementation Details.}
We split each sub-dataset into training and testing sets using a 9:1 ratio. Unlike HOI-Diff~\cite{peng2023hoi}, which computes the Mean Squared Error (MSE) for all motion representations simultaneously, we calculate the loss for the human marker representation and the object motion representation separately. To balance these components, we assign the object motion loss and the contact label loss weights of 0.9 relative to the human motion loss. During training, the model generates 300 frames per sequence—longer sequences are cropped, shorter ones are zero-padded, and padded regions are masked during loss calculation. All experiments were conducted on a single NVIDIA A40 GPU over eight days.

Below, we describe how we apply guidance during the diffusion process, corresponding to the fourth model variant introduced in Sec.~\ref{sec:language}. Our key insight is to perform an additional calculation of object motion in a relative coordinates with respect to markers. By comparing the discrepancy \( L \) between the directly regressed object coordinates and those motion derived from human motion markers, we compute gradients to guide the object trajectories.
The loss function \( L \) is weighted inversely by the distance between each marker and the object to emphasize the influence of closer body parts. We then compute the gradients of this loss with respect to the object's translation \( \boldsymbol{o} \) and rotation \( \boldsymbol{r} \), and the predicted means are updated during the final 30 iterations of diffusion denoising given 1000 iterations in total:
\[
\boldsymbol{\hat{o}} = \boldsymbol{\hat{o}} - \tau_1 \frac{\partial L}{\partial \boldsymbol{o}}, \quad 
\boldsymbol{\hat{r}} = \boldsymbol{\hat{r}} - \tau_2 \frac{\partial L}{\partial \boldsymbol{r}},
\]
where \( \tau_1 = 0.1 \), \( \tau_2 = 0.2 \) is selected.
This guidance directs the diffusion model to generate object motions that consider the relationship with the human. Similar design choices can be found in~\cite{xu2023interdiff,diller2023cg,peng2023hoi,wu2024thor}.

\subsection{Action-Conditioned Interaction Generation} 

\noindent\textbf{Additional Implementation Details.}
We rephrase the action label as ``A person [action] the [object name]'' and treat it similarly to text-conditioned generation, utilizing our interaction-aware text encoder to encode the text as introduced in Sec.~\ref{sec:language} of the main paper. We uses the same data split,  motion representation, and loss functions as the text-to-interaction task. Each experiment is conducted on a single NVIDIA A40 GPU over a duration of eight days. 
\begin{table*}[h!]
\centering
\begin{tabular}{|p{1.5cm}<{\centering}|p{10cm}<{\centering}|p{4cm}<{\centering}|} 
\hline
\textbf{Dataset} & \textbf{Training Objects} & \textbf{Test Objects} \\ \hline
behave & chairwood, keyboard, tablesquare, yogamat, boxmedium, suitcase, basketball, boxsmall, backpack, boxtiny, plasticcontainer, monitor, boxlong, stool, toolbox, chairblack & trashbin, boxlarge, tablesmall, yogaball \\ \hline
chairs & 110, 162, 75, 64, 181, 156, 123, 111, 81, 45, 116, 33, 68, 60, 43, 130, 176, 158, 48, 59, 166, 96, 30, 87, 141, 44, 36, 103, 147, 149, 83, 154, 99, 104, 98, 85, 152, 180, 172, 109, 131, 157, 117, 92, 46, 151, 142, 49, 26, 29, 118, 171, 173, 168, 143, 121 & 15, 17, 24, 25 \\ \hline
grab & toothpaste, spheremedium, cubesmall, table, cylindermedium, cubemedium, cubelarge, apple, duck, cubemiddle, wristwatch, waterbottle, flute, pyramidmedium, piggybank, banana, spheresmall, pyramidsmall, eyeglasses, coffeemug, cylindersmall, torussmall, flashlight, knife, stanfordbunny, pyramidlarge, rubberduck, camera, alarmclock, bowl, wineglass, headphones, cylinderlarge, hammer, stamp, torusmedium, toruslarge, hand, toothbrush, watch, doorknob, body, stapler, train, scissors, mug, elephant, lightbulb, fryingpan, gamecontroller, binoculars, airplane, mouse & phone, cup, teapot, spherelarge \\ \hline
imhd &  broom, pan, baseball, dumbbell, kettlebell, suitcase & chair, skateboard, golf, tennis \\ \hline
intercap & skateboard, stool, racket, soccerball, fantabottle, suitcase & chair, toolbox, umbrella, cup \\ \hline
neuraldome & pillow, smallsofa, trolleycase, monitor, tennis, baseball, flower, chair, pan, case, table, badminton, pingpong, book, keyboard, trashcan, pink & bigsofa, box, talltable, desk \\ \hline
omomo & floorlamp, vacuum\_bottom, mop\_top, largetable, largebox, smallbox, vacuum, monitor, mop\_bottom, plasticbox, vacuum\_top, clothesstand, trashcan, woodchair & smalltable, whitechair, suitcase, tripod, mop \\ \hline
\end{tabular}
\caption{Data split for the task of object-conditioned human generation.}
\label{tab:o2h_split}
\end{table*}

\subsection{Object-Conditioned Human Generation}

\textbf{Additional Implementation Details.} We follow the dataset splitting strategy proposed in OMOMO~\cite{li2023object}. Specifically, we divide HOI interactions into disjoint training and testing set based on object categories. As detailed in Table.~\ref{tab:o2h_split}, the training set includes 168 objects, while the testing set consists of 29 unseen objects. Compared to baseline models, our multi-task model separately computes the human marker reconstruction loss and an additional feature reconstruction loss. The weight for the additional feature reconstruction loss is set to half that of the human marker reconstruction loss. We adopt the same architectural design as the 1-stage and 2-stage baselines in \cite{li2023object}. Our multi-task model consists of four self-attention blocks, each with four attention heads. The dimensions for keys, queries, and values are all set to 256, and each layer produces a 512-dimensional output. The bps feature is configured with a dimension of 256. For training, we use a batch size of 64 over 300k iterations. All experiments are performed on a single NVIDIA A40 GPU, and training the multi-task model, the 1-stage baseline, and both components of the 2-stage baseline takes roughly 23 hours.

\subsection{Human-Conditioned Object Generation}

\textbf{Additional Implementation Details.} We split the dataset into training and testing sets using a 10:1 ratio. In our multi-task model, we define $\boldsymbol{\eta}$ as the distance between human markers and the object. Following the object-conditioned human generation setup, the additional feature reconstruction loss is weighted at 0.5 times that of the human marker reconstruction loss. Our model leverages the same transformer architecture and diffusion framework as the object-conditioned human generation model. For training, we use a batch size of 64 for 260k iterations, with each experiment running on a single NVIDIA A40 GPU and taking approximately 19 hours.

\subsection{Interaction Prediction}

\noindent\textbf{Addtional Implementation Details.}
We use the same transformer architecture and diffusion framework as InterDiff~\cite{xu2023interdiff}.
We substitute the original SMPL representation in InterDiff with marker positions and adjust the corresponding loss function accordingly. As a result, the input and output representations now include marker positions, object angles, object translations, with object geometry as conditions for the diffusion process. Compared to InterDiff, We increase the loss weight for marker MSE by a factor of ten compared to the SMPL MSE loss. For both the InterAct and BEHAVE datasets, we split the data, using 90\% for training and the remaining 10\% of the BEHAVE dataset for testing. For each experiment, the training process is conducted on a single NVIDIA A40 GPU over a span of two days.

\section{License} \label{sec:license}
All data are shared under the CC BY-NC-SA (Attribution-NonCommercial-ShareAlike) license. We will also establish a GitHub repository \url{https://github.com/wzyabcas/InterAct} to receive user feedback on any annotation errors. Users must review and follow the original licenses for each sub-dataset. Please find the licenses of corresponding assets in the code directories, and below is a summary of the licenses for the assets we have used:
\begin{enumerate}
    \item GRAB~\cite{taheri2020grab} uses Software Copyright License for non-commercial scientific research purposes
    \item BEHAVE~\cite{bhatnagar22behave} uses Software Copyright License for non-commercial scientific research purposes
    \item InterCap~\cite{huang2022intercap} uses Software Copyright License for non-commercial scientific research purposes
    \item Chairs~\cite{jiang2022chairs} does not indicate their license but requires to follow the license of each sub-modules they used
    \item HODome~\cite{zhang2023neuraldome} uses Apache License
    \item OMOMO~\cite{li2023object} does not indicate their license
    \item IMHD~\cite{zhao2023im} uses Dataset Copyright License for Non-commercial Scientific Research Purposes
\end{enumerate}

\section{Discussion}\label{sec:ethics}
\noindent\textbf{Limitations.}
Despite that our dataset significantly expands the number of objects to 217 -- nearly ten times more than existing HOI datasets -- we acknowledge that it still possesses scale limitations. While our dataset advances the field by providing enriched annotations and supporting new tasks like text-to-HOI and action-to-HOI, it does not cover the full diversity of in-the-wild object categories encountered in real-world interactions. Although our experiments demonstrate that models trained on our dataset exhibit generalization to out-of-distribution objects within the dataset's scope, as presented in Table~\ref{tab:o2h_after} and our webpage, achieving robust generalization to a broader range of unseen objects will require further expansion. Therefore, while our work makes a significant step toward larger-scale datasets, future efforts are needed to overcome these scale limitations and enhance model performance.

Another limitation of our method lies in the inherent challenges of denoising and correcting full-body HOI data, especially when dealing with significant noise in the original datasets. Issues like floating objects often stem from errors in the initial data rather than shortcomings of our approach. The severity of these errors can be too great for our correction process to handle effectively with the current hyper-parameter configuration, as large distances between the human and object may be identified as no contact and thus remain uncorrected. While adjusting hyper-parameters, such as employing a more aggressive contact threshold, can resolve specific issues like floating objects, we choose to apply unified hyper-parameters across all data for simplicity. Consequently, our method may not correct all artifacts in every scenario. Despite this, it effectively reduces noticeable noise and penetration issues, significantly improving the overall quality and plausibility of the motion data compared to the original.

\noindent\textbf{Ethics Discussion and Potential Negative Societal Impact.}
We collect data on real behavioral information, which could potentially raise privacy concerns. And our correction and augmentation could be used to generate fake data and misleading information. However, we ensure that all collected and generated human data are processed into a format using SMPL~\cite{loper2015smpl} or markers, which significantly reduces identifying details compared to the raw data or images from their original own data. This processed representation effectively enhances privacy. Additionally, all annotations are gathered by the authors  with participant consent. We meticulously review all augmented annotations generated by the language model to ensure they do not contain any harmful information or breach privacy.

\end{document}